\documentclass[conference]{IEEEtran}
\IEEEoverridecommandlockouts
\usepackage{cite}
\usepackage{amsmath,amssymb,amsfonts}
\usepackage{algorithmic}
\usepackage{graphicx}
\usepackage{textcomp}
\usepackage{subcaption}
\usepackage[labelformat=parens,labelsep=quad,skip=3pt]{caption}
\usepackage{graphicx}
\usepackage{xcolor}
\usepackage{comment}
\usepackage{url}
\usepackage{ragged2e}

\usepackage{multirow}
\def\BibTeX{{\rm B\kern-.05em{\sc i\kern-.025em b}\kern-.08em
    T\kern-.1667em\lower.7ex\hbox{E}\kern-.125emX}}
 
\begin{document}

\title{On the combined effect of class imbalance and concept complexity in deep learning
}

\makeatletter
\newcommand{\linebreakand}{%
  \end{@IEEEauthorhalign}
  \hfill\mbox{}\par
  \mbox{}\hfill\begin{@IEEEauthorhalign}
}
\makeatother

\author{\IEEEauthorblockN{Kushankur Ghosh}
 \IEEEauthorblockA{\textit{Department of Computing Science} \\
 \textit{University of Alberta}\\
 Edmonton, Canada \\
 kushanku@ualberta.ca}
 \and
 \IEEEauthorblockN{Colin Bellinger}
 \IEEEauthorblockA{\textit{Digital Technologies} \\
 \textit{National Research Council of Canada}\\
 Ottawa Canada \\
 colin.bellinger@nrc-cnrc.gc.ca}
 \and
 \IEEEauthorblockN{Roberto Corizzo}
 \IEEEauthorblockA{\textit{Department of Computer Science} \\
 \textit{American University}\\
 Washington, DC, USA \\
 rcorizzo@american.edu}
 \linebreakand
 \IEEEauthorblockN{Bartosz Krawczyk}
 \IEEEauthorblockA{\textit{Department of Computer Science} \\
 \textit{Virginia Commonwealth University}\\
 Richmond, VA, USA \\
 bkrawczyk@vcu.edu}
 \and
 \IEEEauthorblockN{Nathalie Japkowicz}
 \IEEEauthorblockA{\textit{Department of Computer Science} \\
 \textit{American University}\\
 Washington, DC, USA \\
 japkowic@american.edu}
}

\maketitle

\begin{abstract}
Structural concept complexity, class overlap, and data scarcity are some of the most important factors influencing the performance of classifiers under class imbalance conditions. When these effects were uncovered in the early 2000s, understandably, the classifiers on which they were demonstrated belonged to the classical rather than Deep Learning categories of approaches. As Deep Learning is gaining ground over classical machine learning and is beginning to be used in critical applied settings, it is important to assess systematically how well they respond to the kind of challenges their classical counterparts have struggled with in the past two decades. The purpose of this paper is to study the behavior of deep learning systems in settings that have previously been deemed challenging to classical machine learning systems to find out whether the depth of the systems is an asset in such settings. The results in both artificial and real-world image datasets (MNIST Fashion, CIFAR-10) show that these settings remain mostly challenging for Deep Learning systems and that deeper architectures seem to help with structural concept complexity but not with overlap challenges in simple artificial domains. Data scarcity is not overcome by deeper layers, either. In the real-world image domains, where overfitting is a greater concern than in the artificial domains, the advantage of deeper architectures is less obvious: while it is observed in certain cases, it is quickly cancelled as models get deeper and perform worse than their shallower counterparts.

\end{abstract}

\begin{IEEEkeywords}
class imbalance, structural concept complexity, class overlap, deep architectures
\end{IEEEkeywords}

\section{Introduction}
This paper presents an in-depth analysis of the impact of structural concept complexity, class overlap and data scarcity on class imbalanced domains in the context of deeper and deeper learning architectures. 
It is well-known that deep architectures have led to successful applications of machine learning in 
domains such as Computer Vision \cite{b5,b6}, Natural Language and Speech Processing \cite{b7,b8} as well as Graph Learning \cite{cao2016deep,zhang2020deep,tian2014learning}. 
Items in such domains can be decomposed in a hierarchical manner and different levels of the hierarchy can be learned by different layers of the Deep learning System which are then combined to assess the identity of the item whose parts they represent \cite{b9,b10}. Despite the indisputable power of deep architectures, they are susceptible to the presence of imbalance among classes, leading to bias toward majority observations. This has given rise to dedicated solutions that make deep models skew-insensitive, via such approaches as instance-based oversampling \cite{Bellinger:2018,Dablain:2021}, generative models \cite{Mullick:2019,Wang:2020}, loss function adaptation \cite{Cao:2019,Wu:2020}, or experience replay \cite{Chrysakis:2020,Korycki:2021}.   
%
In learning from imbalanced data with classical models the issue of challenging domain characteristics, such as instance-level difficulties, play a crucial role \cite{Krawczyk:2016}. Imbalance ratio is not the sole source of learning difficulty and various other factors embedded in the nature of data are negatively impacting the training procedure \cite{Stefanowski:2016}. 

\smallskip
\noindent {\bf Research goal.} The question raised in this paper is whether, in addition to their role in representational learning, deep layers are also beneficial to the handling of challenging domain characteristics. 
We are interested in 
whether increasing the depth of a neural network can help mitigate the challenge caused by class imbalances and their aggravation due to structural concept complexity, class overlap, and data scarcity.

\smallskip
\noindent {\bf Contributions.} In order to address this question, we revisit the simple artificial domains of two early class imbalance studies \cite{b1,b2} that reduce the issues to the bare minimum, thus, allowing us 
to ignore, for a moment, the uncontrollable effects at play in real-world domains as well any representational challenges.  
These experiments allow us to observe how effective deep models are at capturing domain complexities 
other than their representational complexity. 
Because real-world domains, however, are inherently more complex and ultimately more interesting, we repeat our experiments 
on a number of image classification problems derived from the MNIST Fashion and CIFAR-10 domains to see if the conclusions drawn from the artificial domains hold in real-world domains, or whether other phenomena should be considered.

In our study on artificial domains, we test Feedforward Multi-Layer Perceptron (MLP) of increasing depth. In the image domain, we test a Convolutional Neural Network of increasing depth outfitted with two MLP (or dense) layers. In order to gain a deep understanding of the networks' responses to the challenges, we restrict this work to the simple binary classification setting and 
ask the following three questions:
\begin{itemize}
    \item {\bf RQ1:} Does the depth of a model affect its performance  on classification problems of increasing {\bf structural concept complexity}, {\bf class imbalance}, and {\bf data scarcity}\footnote{These data characteristics will be clearly defined in the next section}? 
    \item {\bf RQ2:} Does the depth of a model affect its performance on problems of increasing {\bf class imbalance} and {\bf overlap}?
    \item {\bf RQ3:} Does the depth of a model affect its performance on problems with difficulties from RQ1 and RQ2 occurring simultaneously?
\end{itemize}
The research questions are first addressed in the context of artificial domains and then tested in an image classification setting.

\section{The impact of Depth on the Class Imbalance Problem as it relates to Structural Concept Complexity and Data Scarcity}

\smallskip
\noindent {\bf Description of the Experimental Setup.} This experiment is designed in order to investigate issue raised in \cite{b1} for classical models. In the formulation of the key question below, it is assumed that a class represents a concept (e.g., dogs), that that concept can be further divided into subconcepts (e.g., Poodles, Huskies, Golden Retrievers, German Sheppards, etc.), and that these subconcepts, while distinct from one another  can be mixed amongst subconcepts of the other class (e.g., Huskies or German Sheppards could be represented by subconcepts intertwined between different subconcepts of wolves since various subtypes of wolves and dogs may look very similar to one another and yet quite different from others). The main question asked here is:

  \begin{justify}
    {\bf RQ1:} \textit{Do increases in the number of hidden layers of deep learning networks help mitigate the harmful effects of \textit{a)} increases in the number and mixing level of class subconcepts; \textit{b)} decreases in these subconcepts' size; and \textit{c)} overall decreases in training set size on the class imbalance problem?}
    \end{justify}
\smallskip
\noindent {\bf Domains.} To create a family of domains appropriate to answer the question,  we followed the approach proposed in \cite{b1} to generate domains that vary according to three dimensions: overall size of the data set (s), structural concept complexity (c), and { degree of balance } between the classes (b). The family of domains created by this approach was shown to reflect some of the main challenges surrounding the class imbalance problem and was, therefore, deemed relevant to apply in the case of the deep learning approaches under consideration in this work. 

125 domains were generated as follows: each domain is uni-dimensional with inputs in the [0, 1] range associated with class 1 (+) or 0 (-). The [0, 1] input range is divided into sub-intervals of the same size, each associated with class value 0 or 1. Contiguous intervals have opposite class values. The complexity level, c, can take values from 1 to 5. Depending on its value, different numbers of sub-intervals are created.
An example of a backbone model is shown in Figure \ref{fig:1}. 
\begin{figure}[h]
    \centering    
    \includegraphics[width=85mm]{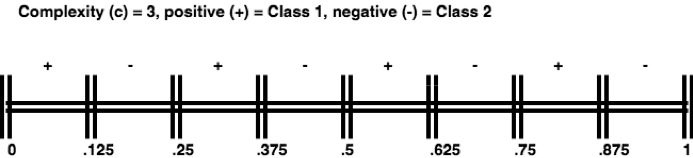}
    \bigskip
    \caption{Domain backbone of Complexity 3. In this one-dimensional family of domains, the complexity of the  task increases as the number of alternating sub-concepts of each class increases.}
    \label{fig:1}
\end{figure}
The backbone in the figure can be understood as representing a two-class domain where each class is composed of 4 subconcepts that are mixed amongst themselves. Maybe, for example, the data located between 0 and .125 represents the subspecies of dogs A which is very closely related to the subspecies of wolves A located between .125 and .25, while the location between .25 and .375 represents subspecies of dogs B, while the location between .375 and .5 represents subspecies of wolves B; and so on. 
%
%

The value of c is used to determine the number of subclusters present in the backbone that ranges within [0,1]. The number of subclusters is calculated as $2^{c}$ and the width of each of these sub-sections is calculated as $\frac{1}{2^{c}}$. As illustrated in Figure \ref{fig:1}, the distribution of Class 1 and Class 0 is determined by assigning them regular, alternating sub-intervals. This is done regardless of the size of the training set or its degree of imbalance. Once the backbone is generated based on the value of c, actual data points
are generated within each sub-interval by generating points at random using a uniform distribution. The number of points sampled from each interval depends on the size of the domain as well as on its degree of balance.

Our investigation revolves around two dataset sizes, which we will refer to as sizes 1 and 5 (or s=1 and s=5), according to \cite{b1}. In \cite{b1}, prior to considering the balance level, b, the total number of examples in the size 1 experiments is calculated as $\left(\frac{5000}{32}\times2\right)$ where each sub-interval contains $\left(\frac{\frac{5000}{32}\times2}{2^{c}}\right)$ examples. In the size 5 experiment, the dataset holds a total number of $\left(\frac{5000}{32}\times2^{5}\right)$  examples with $\left(\frac{\frac{5000}{32}\times2^{5}}{2^{c}}\right)$ instances in each of the sub-intervals. 

Once the basic number of instances per sub-interval is determined, we decrease that number for class 0, the minority class, according to the degree of balance, b. Meanwhile, the number of instances in the Class 1 sub-intervals representing the majority class remain the same as discussed in the previous paragraph. The number of instances belonging to the Class 0 sub-intervals is calculated as $\left(\frac{\left(\frac{\frac{5000}{32}\times2}{2^{c}}\right)}{\frac{32}{2^{b}}}\right)$ for size 1 and $\left(\frac{\left(\frac{\frac{5000}{32}\times2^{5}}{2^{c}}\right)}{\frac{32}{2^{b}}}\right)$ for size 5. The expression $\left(\frac{32}{2^{b}}\right)$ gives a limit of 5 to the degree of balance in our experiment. When b=5, the number of instances in each of the sub-intervals is the same and the data set is perfectly balanced. This states that the value of b is inversely proportional to the disparity or the degree of imbalance between the classes.  

\smallskip
\noindent {\bf Train/Test regimen.} We conducted two separate series of experiments 
1) 10 Fold Stratified Cross-Validation approach and 2) Balanced Testing approach. For the Stratified Cross-Validation approach we observed that, in certain scenarios when the degree of balance is low, in each folds the testing set holds a very smaller number of examples belonging to Class 0. This makes it difficult to assess the performance of the models. To overcome this problem we used a technique which we call the balanced testing approach by forming a testing dataset containing 1000 examples in each of the sub-clusters of the proposed backbone. This not only provides an unbiased testing framework, but also provides a wide variety of data to test on. This approach, therefore, helps understand the actual potential of the classification models. We report the Balanced Testing results in the paper and include the stratified cross-validated results in the supplementary material section.

Although we calculated our results based on a variety of metrics, we decided to report the geometric mean (G-Mean) after noticing that other metrics such as the F1-measure and the balanced average all lead to the same conclusions. These results are available upon request. Given the imbalanced nature of our datasets, we understandably opted against the micro average reporting regimen. The macro and the weighted regimen on the other hand gave us the same results given that our balanced testing approach uses as many instances of the minority and majority class for testing.{ A formal explanation of why this is happening is presented in the supplementary material.} The same train/test regimen was used for all the experiments presented in this paper.

\begin{figure*}[t]
  \centering
  \begin{subfigure}{3cm}
    \centering\includegraphics[width=3cm]{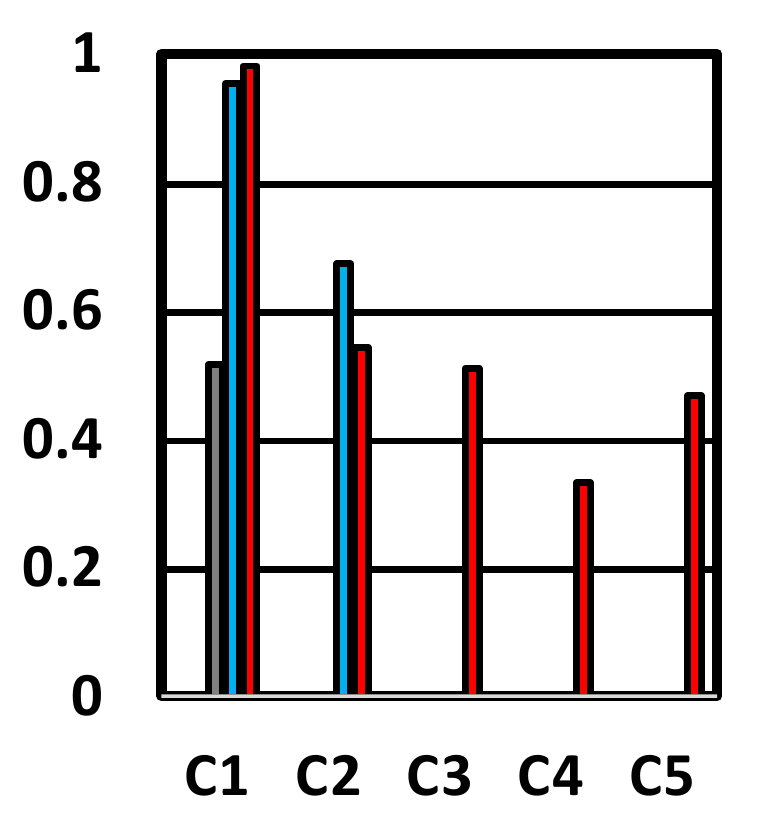}
    \caption{Model 1}
  \end{subfigure}
  \begin{subfigure}{3cm}
    \centering\includegraphics[width=3cm]{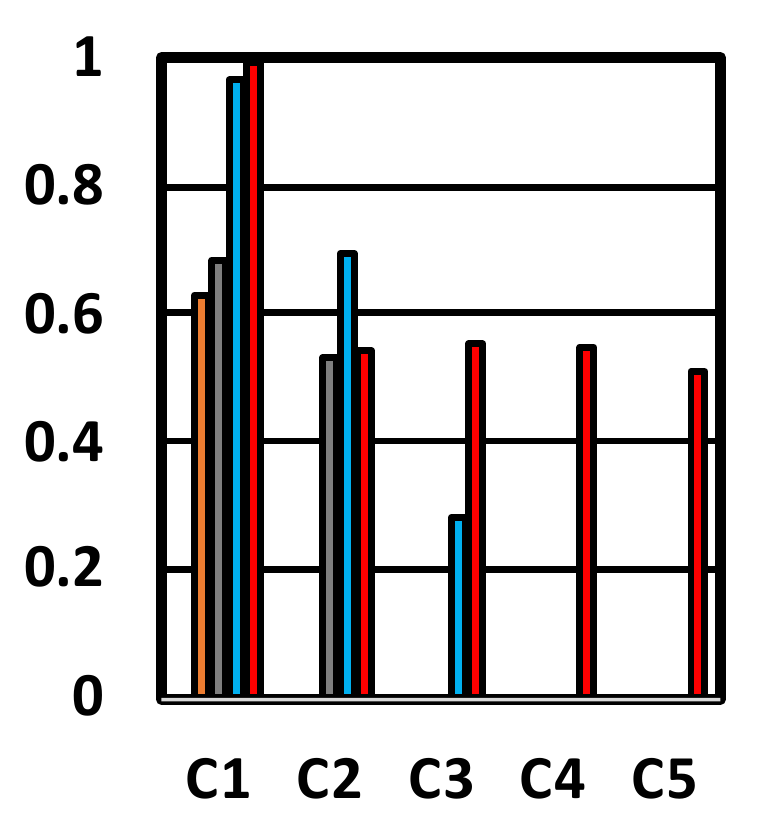}
    \caption{Model 2}
  \end{subfigure}
  \begin{subfigure}{3cm}
    \centering\includegraphics[width=3cm]{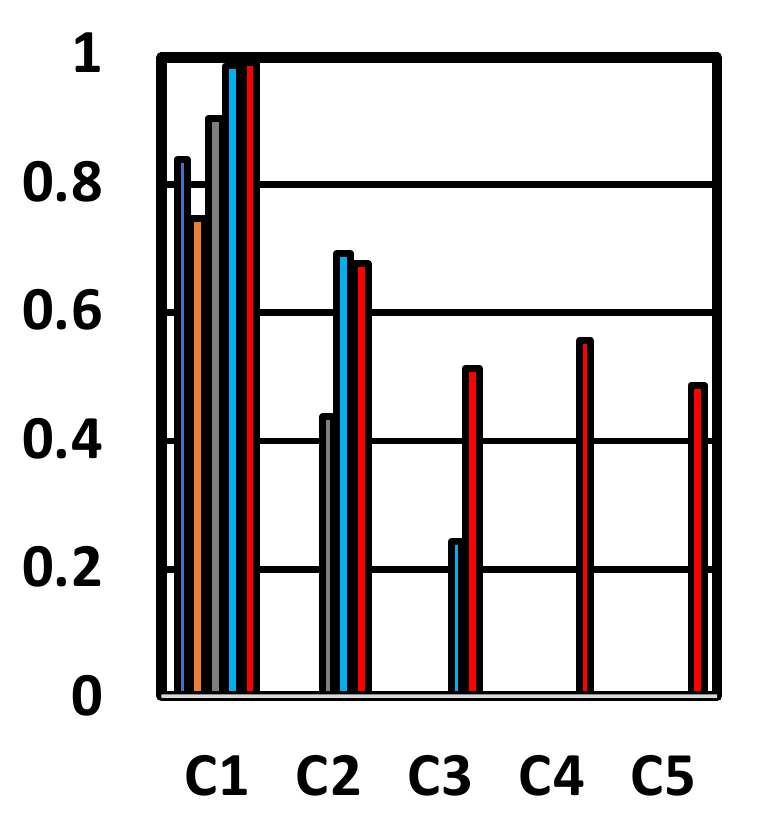}
    \caption{Model 3}
  \end{subfigure}
  \begin{subfigure}{3cm}
    \centering\includegraphics[width=3cm]{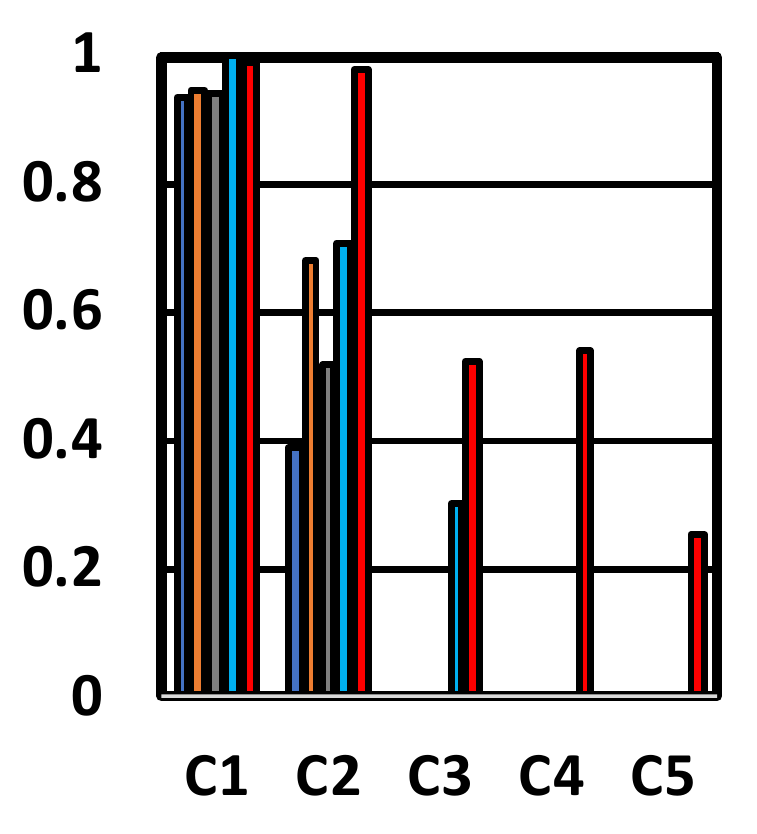}
    \caption{Model 4}
  \end{subfigure}
  \begin{subfigure}{3cm}
    \centering\includegraphics[width=3cm]{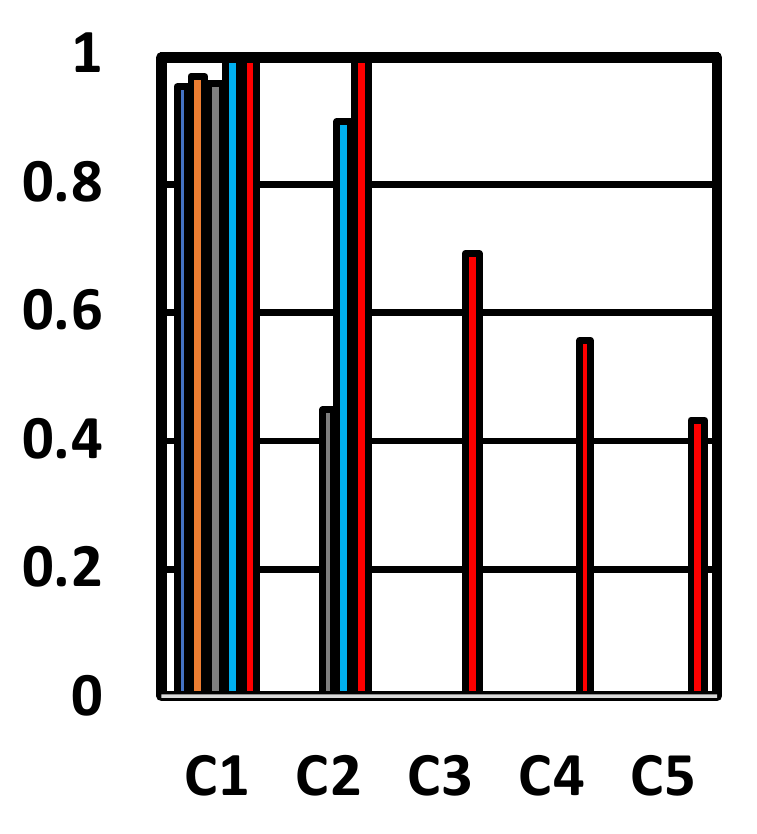}
    \caption{Model 5}
  \end{subfigure}
  \caption{MLP generated Macro G-Mean Scores by doing Balanced Testing for Size 1: (a) Model 1, (b) Model 2, (c) Model 3, (d) Model 4, and (e) Model 5. These plots show that, in sparse data conditions, as the structural complexity of the problem increases, so does the impact of the class imbalance problem. Deeper layers are able to help but do not overcome the problem in the more complex and imbalanced domains.
  }
  \label{fig:9}
\end{figure*}

\begin{figure*}[t]
  \centering
  \begin{subfigure}{3cm}
    \centering\includegraphics[width=3cm]{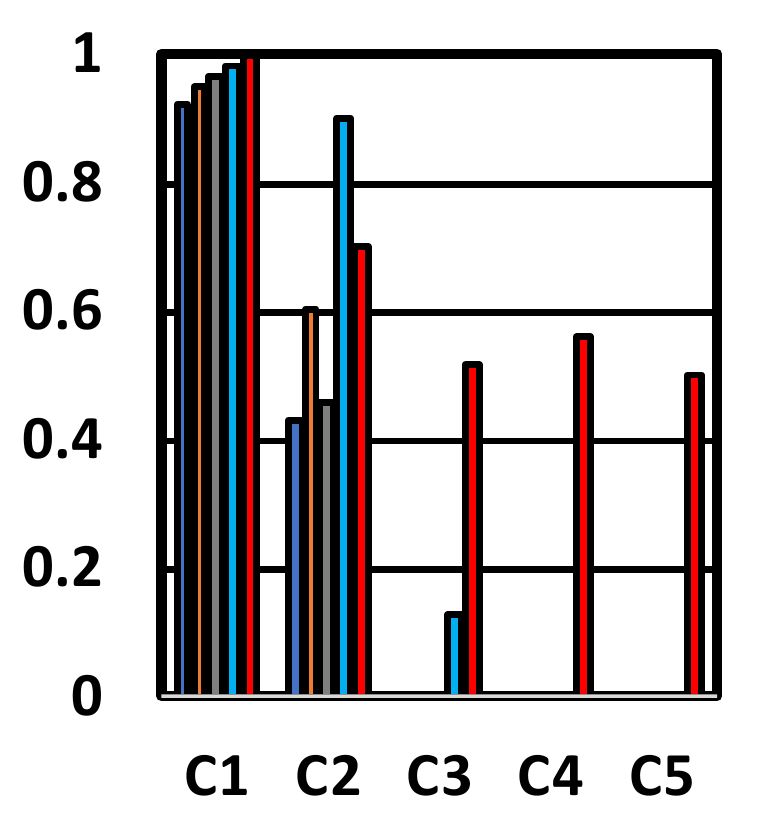}
    \caption{Model 1}
  \end{subfigure}
  \begin{subfigure}{3cm}
    \centering\includegraphics[width=3cm]{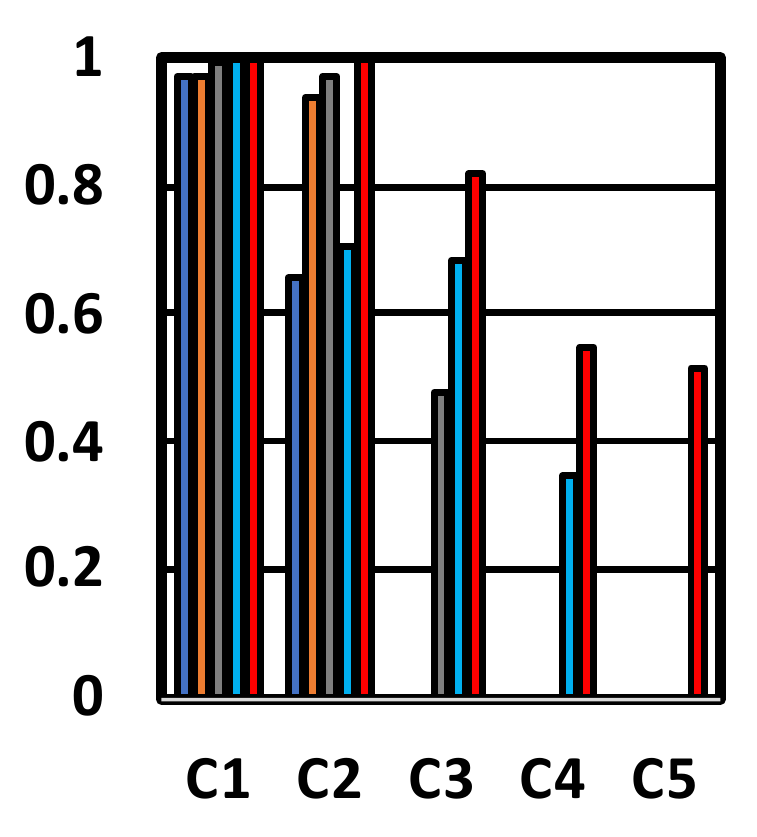}
    \caption{Model 2}
  \end{subfigure}
  \begin{subfigure}{3cm}
    \centering\includegraphics[width=3cm]{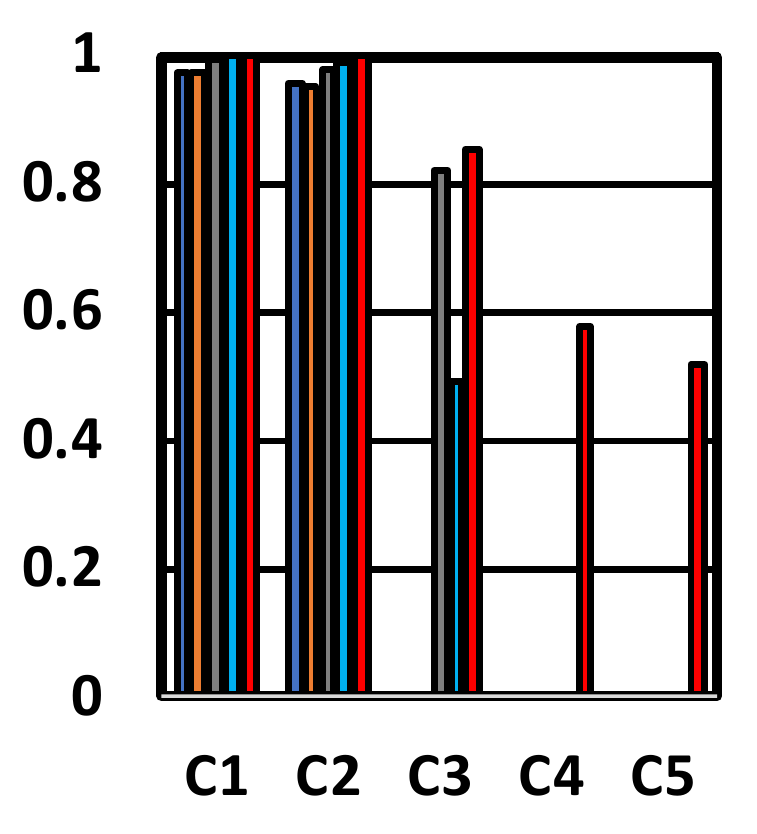}
    \caption{Model 3}
  \end{subfigure}
  \begin{subfigure}{3cm}
    \centering\includegraphics[width=3cm]{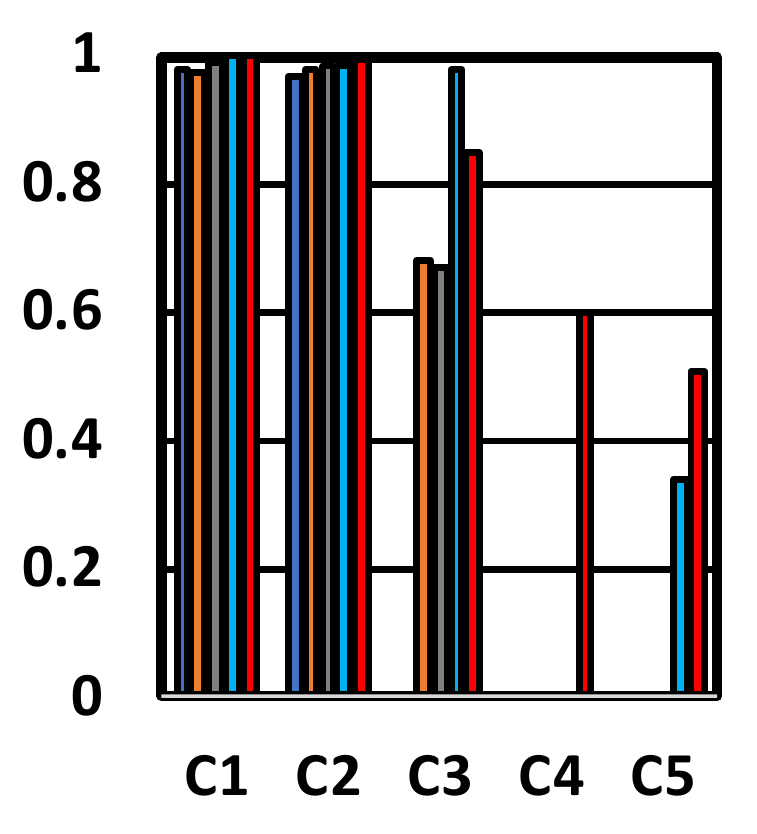}
    \caption{Model 4}
  \end{subfigure}
  \begin{subfigure}{3cm}
    \centering\includegraphics[width=3cm]{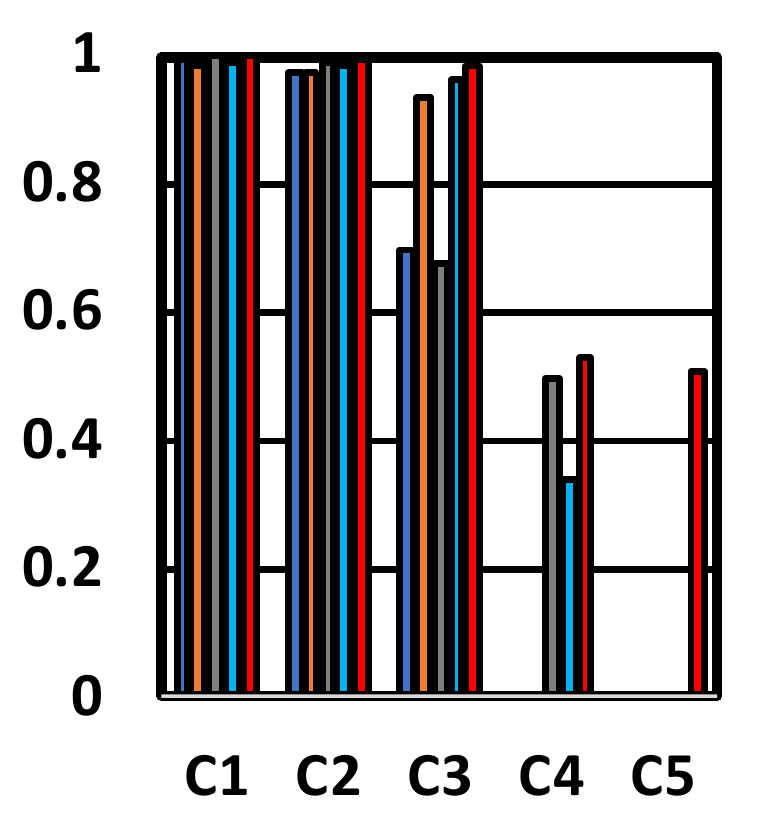}
    \caption{Model 5}
  \end{subfigure}
  \caption{MLP generated Macro G-Mean Scores by doing Balanced Testing for Size 5: (a) Model 1, (b) Model 2, (c) Model 3, (d) Model 4, and (e) Model 5. These plots show results similar to those obtained for size=1, but the issues are less pronounced, which suggests that large data sets along with deep models mitigate the issues caused by concept complexity in class imbalance settings. Nonetheless, deep layers are not necessary able to fully overcome the problem.
  }
  \label{fig:10}
\end{figure*}

\smallskip
\noindent {\bf Models and their parameters.} The experiments for this section, as well as for those of sections III and IV are conducted on five different depth of Feedforward Networks or Multi-Layer Perceptron (MLP) 
to show how the linear effect of increasing the depth of MLPs affects classification. 
Each of the MLP models are termed Model-x where x stands for the number of hidden layers and takes a value between 1 and 5. 
We start from the shallowest model (Model-1) and reach up to the deepest model with 5 hidden layers (Model-5). For each of the networks, we report the optimal results recorded after experimenting with 2, 4, 8, and 16 Hidden Units (HU) in each layer. We trained each of the MLP networks for 300 epochs, with a learning rate of 0.001, using the Adam optimizer. 


\smallskip
\noindent {\bf Results.} The results for this experiment are reported in Figures \ref{fig:9} and \ref{fig:10}
which report the results obtained for sizes s=1 and s=5, respectively. Each figure reports 5 sets of barplots labeled from (a) to (e) that each show the G-Mean  obtained at a different depth, going from one hidden layer (plot (a)) to 5 (plot (e)). Within each plot, 5 clusters of bars are shown. The cluster of bars labeled c1 represents the easiest (linearly separable) problem with a single concept/subconcept per class; the second, c2, represents the second easiest with two (alternating) subconcepts per class, until c5, the most complex with 16 (alternating) subconcepts per class. Within each cluster are 5 bars (not always visible if their value is 0). The bar furthest to the left represents the smallest degree of balance (or highest degree of imbalance) b=1; while the rightmost bar represents the balanced situation b=5. Figure \ref{fig:9} reports the results obtained for datasets of small size (s=1), while Figure \ref{fig:10} reports the results obtained for datasets of large size (s=5). 

Our first major observation, which serves as a prelude to the main question we seek to answer in this section, is that like in the classical classifier case, as the degrees of domain complexity and class imbalances increase, the performance of the classifiers decreases. This is seen in each subplot, but we choose to illustrate it on Figure~\ref{fig:9} (c) which is representative of the tendency we observe. In that subplot, the c1 cluster generally obtains a high G-Mean value (around .7 or .8 for the two leftmost (most imbalanced) bars), with the rightmost bar obtaining a perfect value of 1.
The c2 cluster has only 3 bars showing since at  this level of complexity, the two most imbalanced levels, b=1 and b=2, obtain a G-Mean of 0. The highest values reached by the rightmost bars at this level of complexity is only of .7. The situation continues to deteriorate as the level of complexity increases. Figure~\ref{fig:10} shows the corresponding graphs for size s=5. By looking at subplot \ref{fig:10} (c), we see clearly that a larger sample size is helpful since decent G-Means are obtained for both degrees of complexity c1 and c2, which was not the case for s=1. It is only at level of complexity c3, that the performance starts deteriorating in the same kind of fashion as it did for s=1.


Our second major observation regards the number of layers used in our networks and thus answers our main question. What we observe is that
 as we add more layers, the model obtains better and better performance. Indeed, comparing, for example, Figure \ref{fig:9} (a) to Figure \ref{fig:9}(e), we see that while in Figure \ref{fig:9}(a), at complexity level c1, only balance levels b=4 and b=5 managed to obtain G-mean results close to 1, such results were obtained for all levels of imbalances for c1 in Figure \ref{fig:9}(e). In fact, the two most balanced levels, b=4 and b=5, for c2 in Figure \ref{fig:9}(e) are also close to 1 whereas, they did not exceed .7 in \ref{fig:9}(a). The improvement is even more dramatic for large overall sample size s=5, in Figure \ref{fig:10}, where all levels of imbalances for concept complexities c1 and c2 are perfectly handled and concept complexity c3 is moderately-to-well handled depending on the balance level of the data. Despite the improvement observed with the addition of layers, it is important to note that 5 layers are nowhere near able to handle complexity levels c4 and c5, let alone when the imbalance is high (e.g., b=1, 2 or 3) or when the data is scarce (s=1). This suggests that MLP Networks' depth alone may not be sufficient to handle class imbalances when the sample size is small, the concept complexity high, and the data highly imbalanced. It may not be sufficient in all cases, even when the sample size is high.
 
 The results obtained using stratified cross-validation {(the results for Model 1 and 5 are presented as supplementary material and the rest is available upon request)} show the same pattern though in an exaggerated manner since the testing set contains fewer and fewer minority examples as the class imbalance increases, as discussed previously.

\section{The impact of Depth on the Class Imbalance Problem as it relates to Class Overlap}

\begin{figure}[t]
  \centering
  \begin{subfigure}{2.7cm}
    \centering\includegraphics[width=2.7cm]{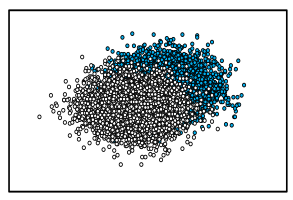}
    \caption{}
  \end{subfigure}
  \begin{subfigure}{2.7cm}
    \centering\includegraphics[width=2.7cm]{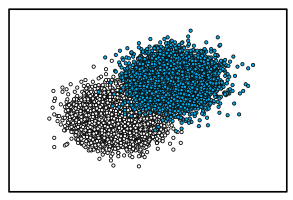}
    \caption{}
  \end{subfigure}
  \begin{subfigure}{2.7cm}
    \centering\includegraphics[width=2.7cm]{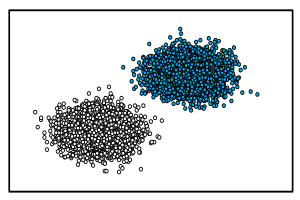}
    \caption{}
  \end{subfigure}
  \caption{Instances of Overlapped Domain: (a) Overlap 3, (b) Overlap 5, and (c) Overlap 9}
  \label{fig:16}
\end{figure}


\smallskip
\noindent {\bf Description of the Experimental Setup.} This experiment is designed in order to investigate issue raised in \cite{b2} for classical models. The main question asked here is:

  \begin{justify}
    {\bf RQ2:} \textit{Do increases in the number of hidden layers of deep learning networks help mitigate the harmful effects of class overlaps on the class imbalance problem?} 
    \end{justify}
To answer this question, we generate datasets with different degrees of overlapping distribution. In this context, class overlap takes the shape displayed in Figure \ref{fig:16} where class overlaps of different degrees were created by moving the means of 
each class closer and closer to each other.
\footnote{Note that the experiments in the previous section also dealt with overlap, but the overlap considered there was highly structured in that within the region of overlap, the overlapping subconcepts occupied distinct regions. In this section, we consider overlap where in the overlapping region, the two classes do not occupy distinct sub-regions. We distinguish the two phenomena by calling the first one ``structural concept complexity" (or simply ``concept complexity") and the second, ``overlap".}


 We conducted the analysis of Feedforward MLP networks of different depth on 10 different overlapped distributions with different degree of imbalance generated as described in the next subsection.


\smallskip
\noindent {\bf Domains.} In order to create overlapping domains, we followed the approach proposed by \cite{b2}. The most complex dataset is such that the two classes overlap completely and the easiest one is completely separable. {Eight other domains were generated in between these two extremes, some instances of which are illustrated in Figure \ref{fig:16} (though in reality, the domain is 5-dimensional rather than 2-dimensional).} The 10 different distributions are entitled Overlap 1 to Overlap 10. These distributions are generated by following a 5-Dimensional Gaussian Distribution with a Standard Deviation of 1. Initially, for Overlap 1 (representing the highest level of overlap) the mean for each class is the same, at 0.5. It is then incremented by 1, step-wise, up to nine times to obtain the 9 other distributions. Each of these distributions, itself, contains 12 separate datasets 
differing in the amount of minority instances. The total amount of data is kept at 10,000 in each of the sets by assigning the quantity of minority samples as 1\%, 2.5\%, 5\%, 10\%, 15\%, 20\%, 25\%, 30\%, 35\%, 40\%, 45\%, and 50\%. With 50\% minority samples in a dataset, we have a balanced set of examples (5000 in the majority and 5000 in the minority). Similar to the testing process proposed for the 
experiments in the previous section, we conducted the evaluation based on 10 Fold stratified cross-validation and Balanced Testing. A total of 10 balanced test sets are designed by following the distribution of Overlap 1 to Overlap 10. Each of the class in the testing set is given a total of 2000 examples to maintain the balance. For consistency, we continue reporting the Balanced Testing results, but the results obtained on the cross-validation experiments show similar patterns {(results for Models 1 and 5 are presented in the supplementary section)}.

\begin{figure}[t]
  \centering
  \begin{subfigure}{8.5cm}
    \centering\includegraphics[height=3cm,width=8.5cm]{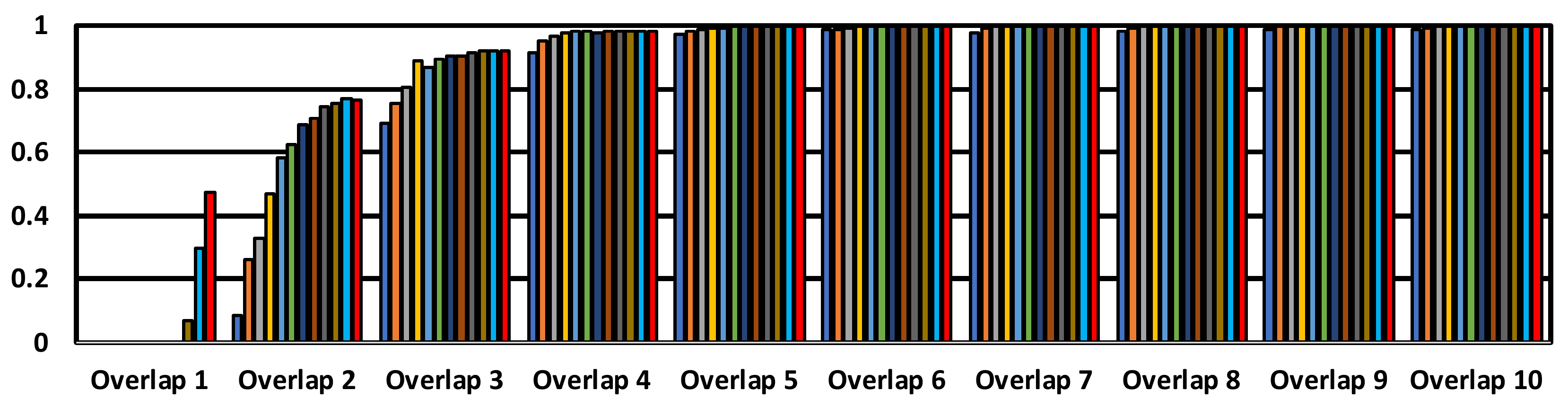}
    \caption{}
  \end{subfigure}
  \begin{subfigure}{8.5cm}
    \centering\includegraphics[height=3cm,width=8.5cm]{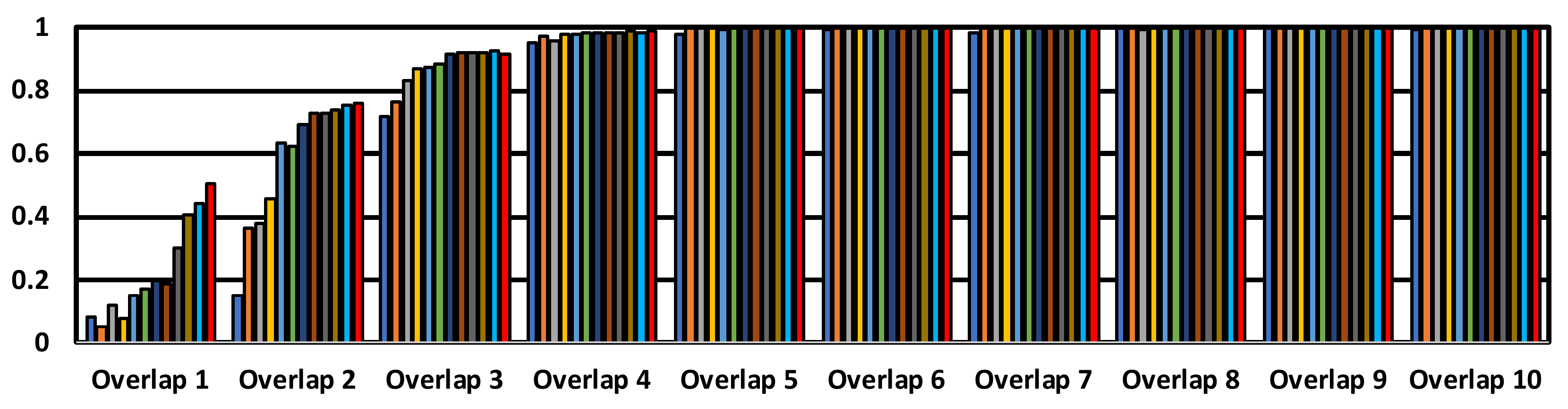}
    \caption{}
  \end{subfigure}
  \caption{MLP generated Macro G-Mean Scores by doing Balanced Testing on the Overlapped Datasets: (a) Model 1, 
  (b) Model 5. 
  The graphs show that increasing the depth of the network is not particularly useful, except, minimally, in Overlap 1 where the distributions are fully overlapped and the problem, therefore, not fully meaningful. 
  }
  \label{fig:19}
\end{figure}


\smallskip
\noindent {\bf Results.} The results obtained in the overlap experiments on the balanced testing set are reported in Figure~\ref{fig:19}. 
Due to space constraints and because the results do not change significantly from layer to layer except in the case of Overlap 1 (when the two classes overlap completely), we only display the results obtained with one hidden layer and those obtained with five hidden layers. In more detail,
each subplot represents a different number of hidden layers: 
1 (subplot (a)) and 5 (subplot (b)). Within each subplot, each cluster of bars represents the results obtained at a different overlap level (from Overlap 1 to Overlap 10). Within each cluster, there are 10 bars, each representing the 12 different imbalance levels previously discussed and ordered from most imbalanced (1\% minority) to completely balanced (50\% minority).


As before, the first major observation serves as a prelude to the main question we seek to answer in this section. The observation regards the effect of overlaps of the kind considered here on the class imbalance problem. Like in the classical case, our results show that, at least for overlap levels 1 to 3, class overlap is very detrimental to MLP Networks and worsens when combined with increasing degrees of class imbalance (Figure \ref{fig:19}). By class overlap 4, however, the effect is minimal and it disappears completely starting with class overlap 5. {It is interesting to map these results to the plots in Figure~\ref{fig:16}  where Figure ~\ref{fig:16} (a) shows the significant overlap encountered in overlap 3 and Figure ~\ref{fig:16}(b) shows the present, but not as significant overlap in overlap 5}. It is remarkable that the MLP Networks actually do so well when encountering overlap level 3 and do perfectly well, irrespective of imbalanced levels in overlap level 5. 


The second major result concerns the main question raised in this section, namely, the effect of increasing the depth of the network architecture. Overall, what we found is that increasing the depth of the networks has minimal effect in the case of overlap. In more detail, the results in Figure \ref{fig:19} show that for overlap level 1, when the two classes are completely overlapped, an increase in depth is beneficial as we see bars that did not appear in the first cluster of subplot (a) now appear (i.e., achieve non-zero G-Mean) and increase in height in subplots (b) to (e) (with only (e) shown here). Given that the two distributions are completely overlapped, though, it is not clear that what the deeper networks learn is particularly meaningful. More significantly, for all the other overlap levels, the increase in depth has no or only a minimal effect.

\section{The impact of Depth on the combined effects of Class Imbalance and Overlapping as it relates to Real World Problems}

While the last two sections sought to study the problems of structural concept complexity and overlap separately, the purpose of this section is to study the two problems simultaneously, in the way they are expected to occur in real-world problems.

\smallskip
\noindent {\bf Description of the Experimental Setup.} Our goal is now to integrate the issues of structural concept complexity, class overlap and class imbalance together. We use the same backbone as the one used in Section II, but in contrast with the experiments of that section, we allow distinct sub-regions of the data to overlap, and consider different degrees of overlap. 

The main question asked here is:

  \begin{justify}
    {\bf RQ3:} \textit{Do increases in the number of hidden layers of deep learning networks help mitigate the harmful effects of learning difficulties from RQ1 and RQ2 occurring simultaneously?}
    \end{justify}


\smallskip
\noindent {\bf Domains.} To generate the integrated domains, we considered a backbone with moderate level of concept complexity (c = 2) and size = 5. To make the experiment manageable, the size and concept complexity are kept constant and only the balance level, b, and overlapping degree, v, are modified. In these domains, we incorporate overlaps between each subconcept of the backbone by generating Gaussian distributions centered in the middle of each sub-interval. 
This time, and unlike in Section III of the paper, the degree of overlap between subconcepts is increased by increasing the variance of each distribution.
We concentrate on 5 different variation of overlaps (v1, v2, v3, v4, v5) represented in Figure \ref{fig:distPlotsGaussianBackbone} that is arranged in order, starting from low overlap (v1) to high overlap (v5). Each of these distributions are divided into 5 levels of b (as we did previously)  by generating $\left(\frac{1250}{\frac{32}{2^{b}}}\right)$ examples in each minority subconcept and a constant number of 1250 examples in the majority subconcepts. We constructed a testing set by following the balanced testing approach used previously and generating 2000 instances for each class (and, more specifically, 1000 instances in each subconcept). 


\smallskip
\noindent {\bf Results.} The results are displayed in Figure \ref{fig:GaussianBackboneMLPGMean} for c2, s=5 and b and v varying between 1 and 5. The same network architecture was used as in the previous sections and each plot represents a different network depth. Within each plot, each cluseter of bars represents a different overlap level, going from no overlap (v1) on the left to highly overlapped (v5) on the right. Within each cluster, as before, b varies from least balanced (b1) to fully balanced (b5). 
 \begin{figure*}[t]
     \centering\includegraphics[scale=0.23]{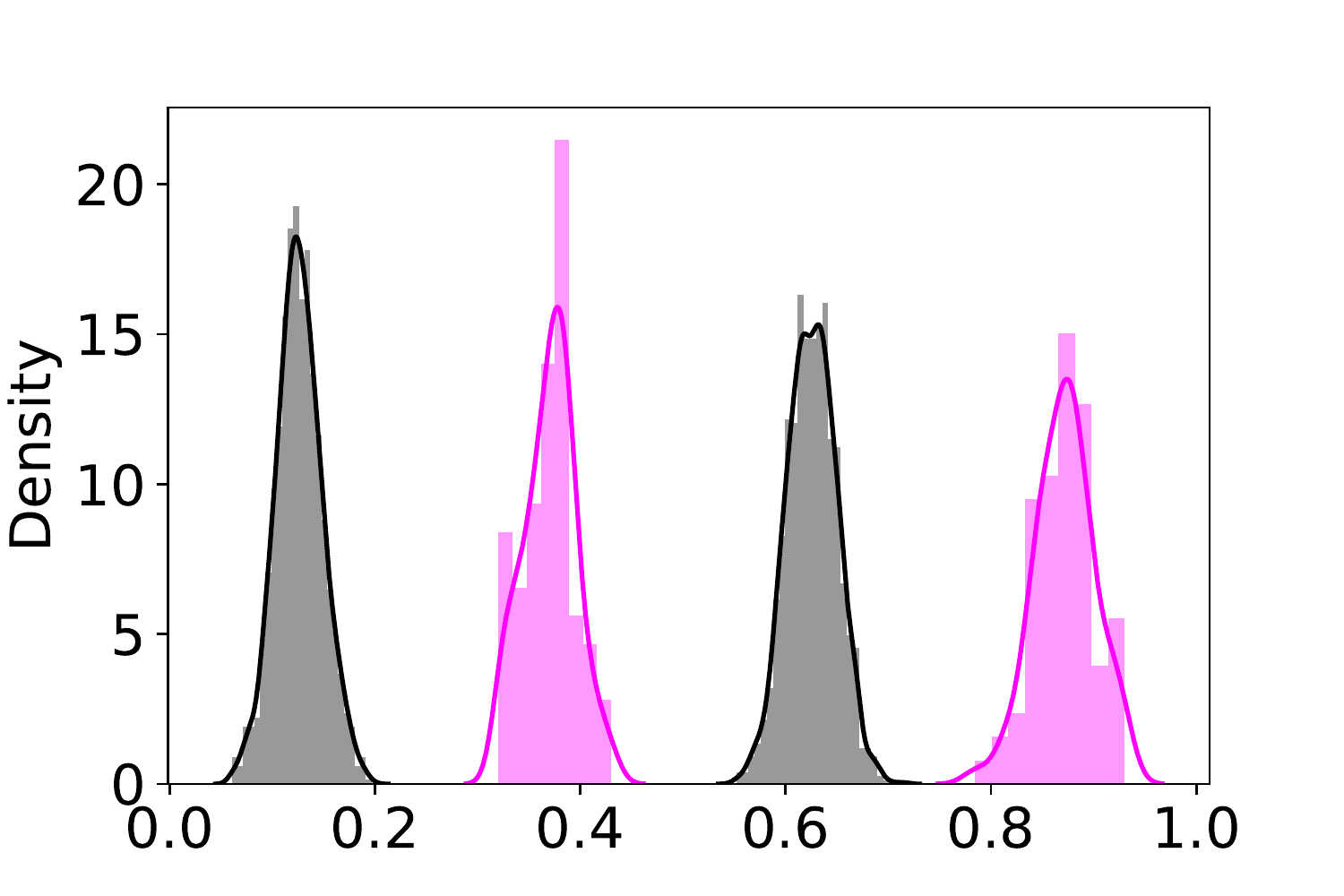}
     \centering\includegraphics[scale=0.23]{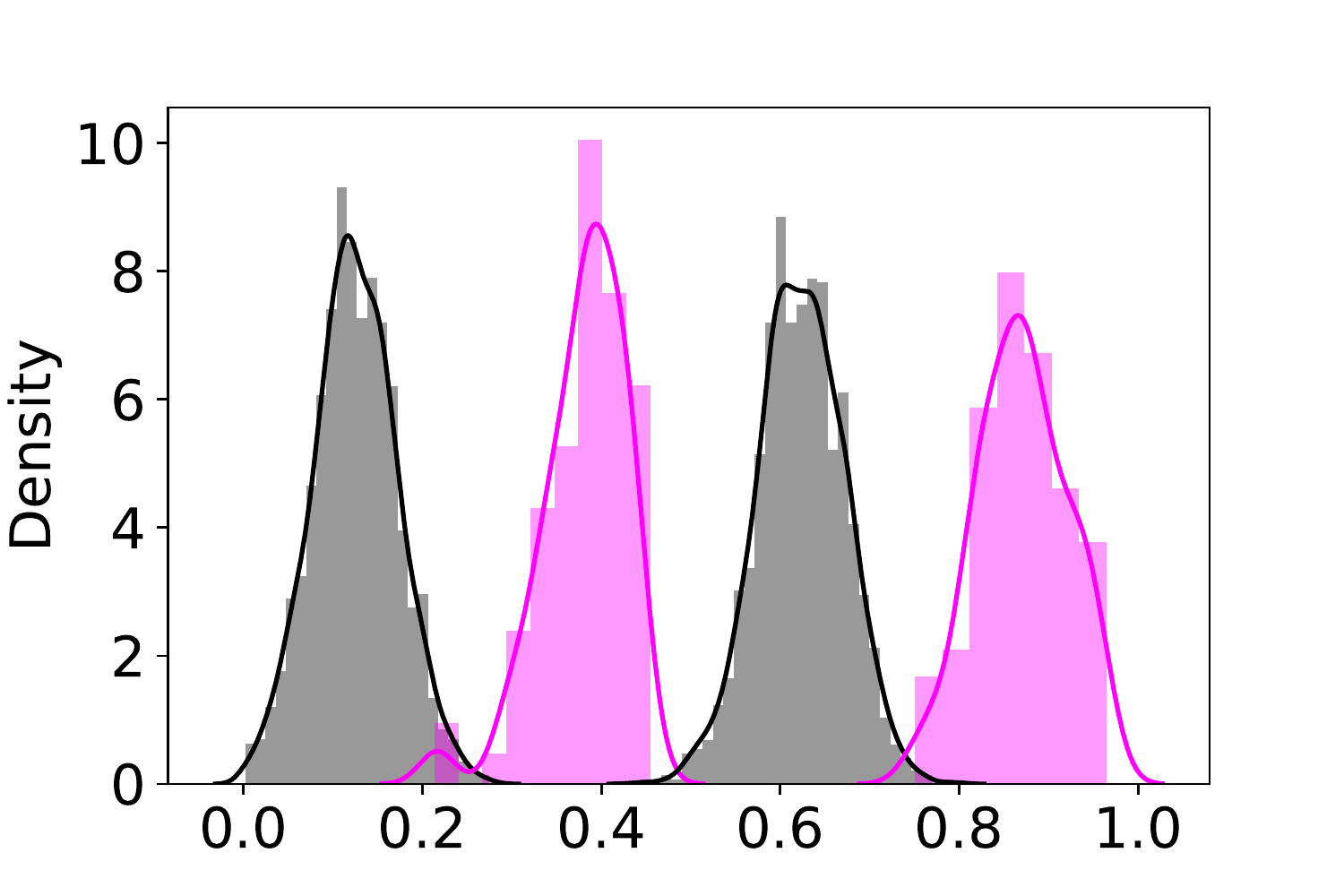}
     \centering\includegraphics[scale=0.23]{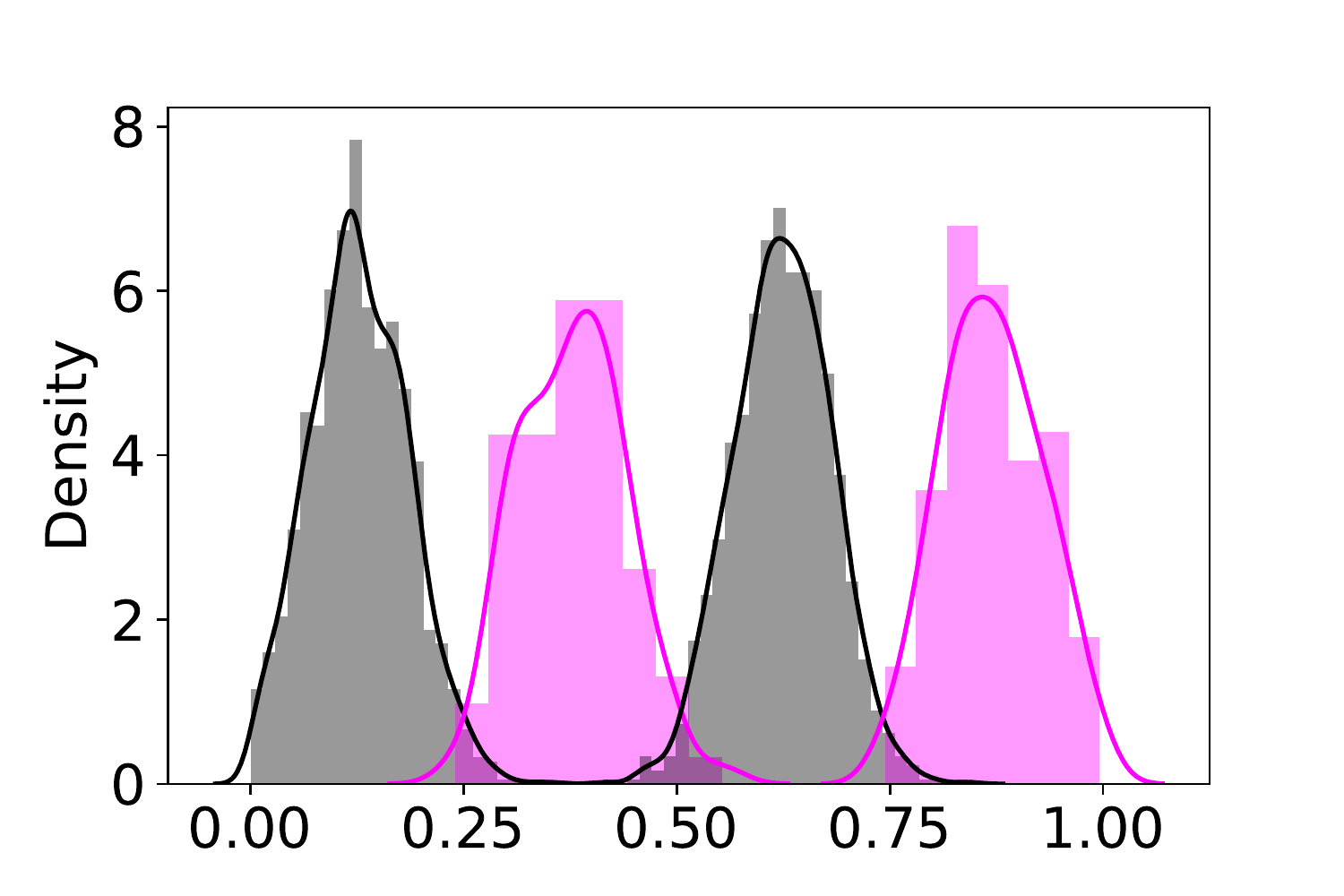}
     \centering\includegraphics[scale=0.23]{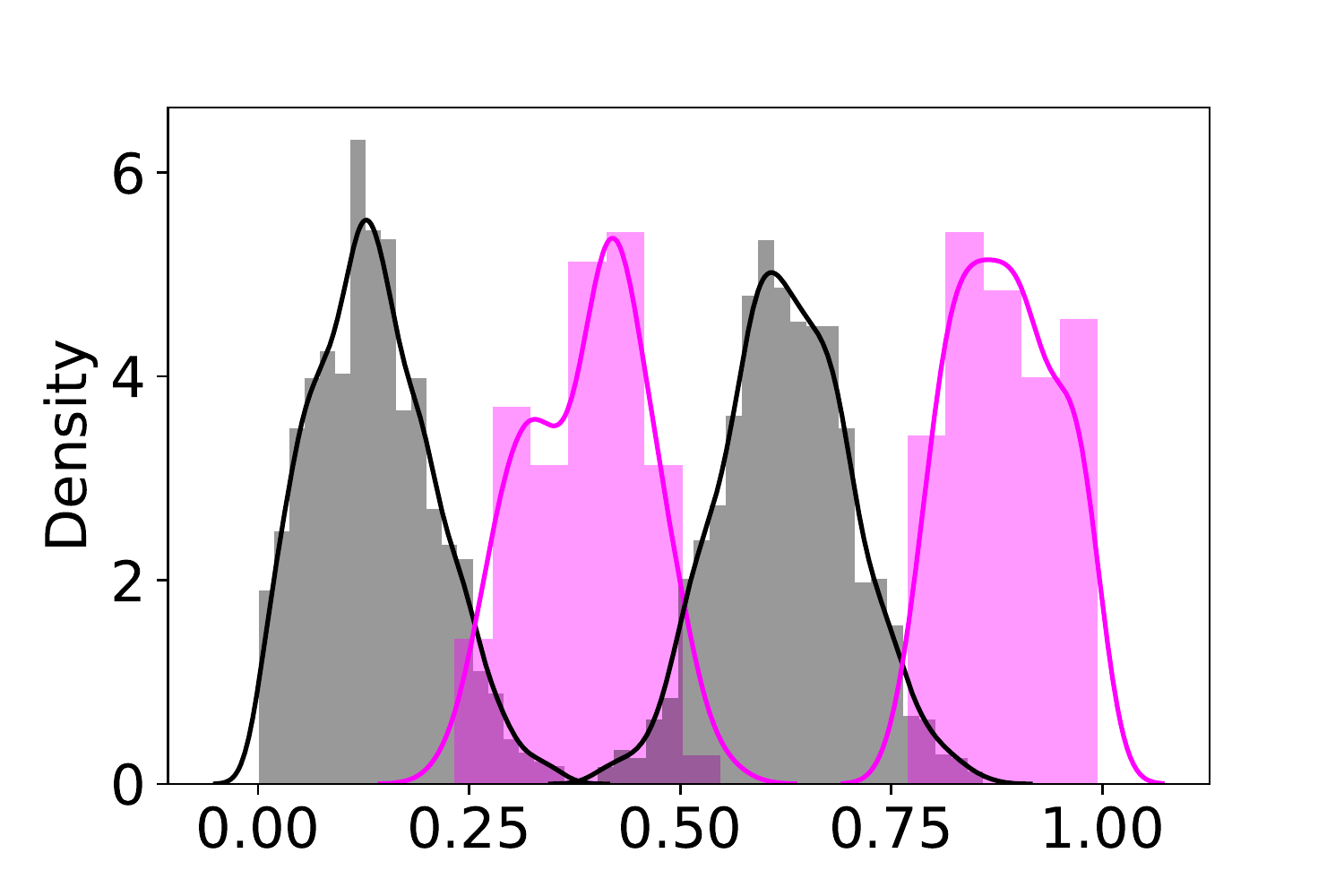}
     \centering\includegraphics[scale=0.23]{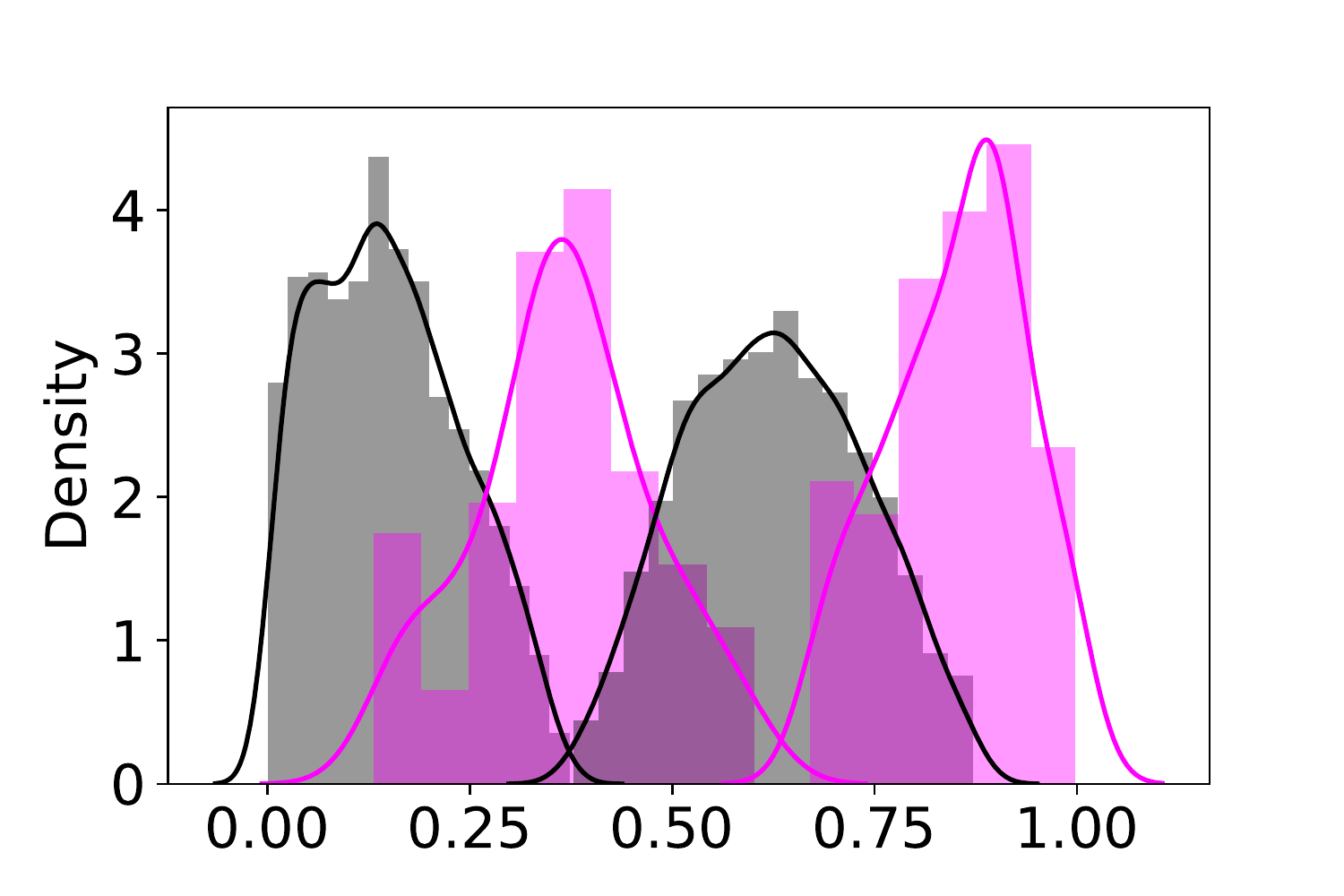}
   \caption{Plots of Overlapped Gaussian Distributions on Backbone sorted according to overlap.}
   \label{fig:distPlotsGaussianBackbone}
 \end{figure*}
 
 \begin{figure*}[t]
  \centering
  \begin{subfigure}{3cm}
    \centering\includegraphics[width=3cm]{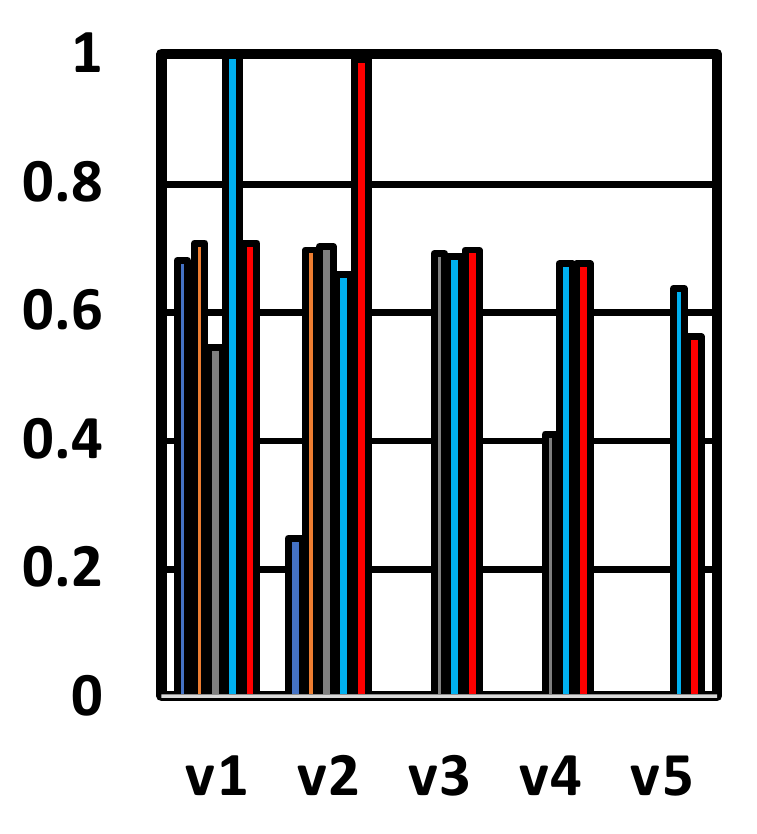}
    \caption{Model Depth 1}
  \end{subfigure}
  \begin{subfigure}{3cm}
    \centering\includegraphics[width=3cm]{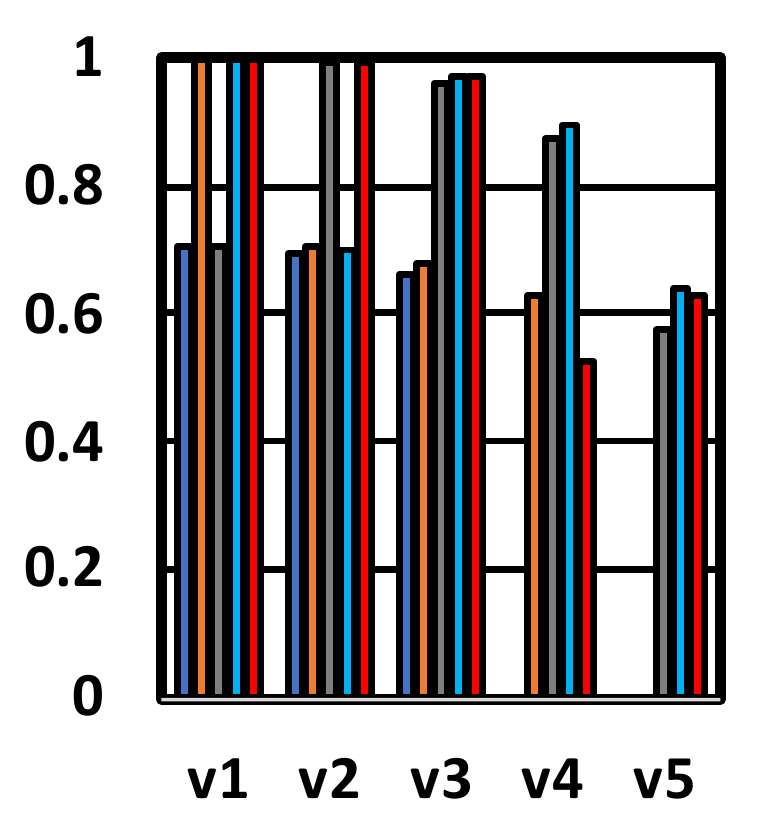}
    \caption{Model Depth 2}
  \end{subfigure}
  \begin{subfigure}{3cm}
    \centering\includegraphics[width=3cm]{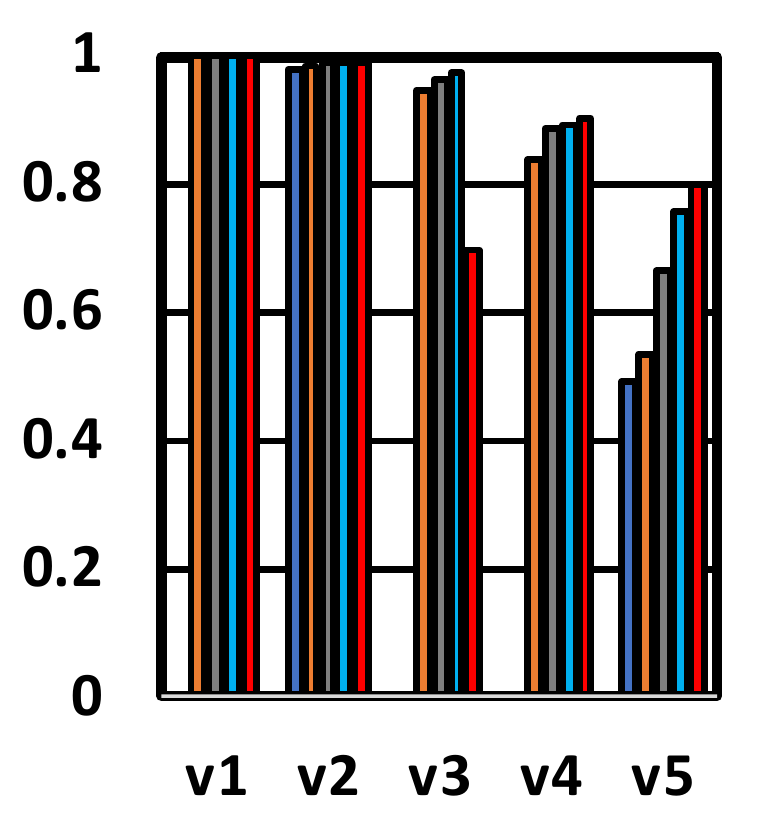}
    \caption{Model Depth 3}
  \end{subfigure}
  \begin{subfigure}{3cm}
    \centering\includegraphics[width=3cm]{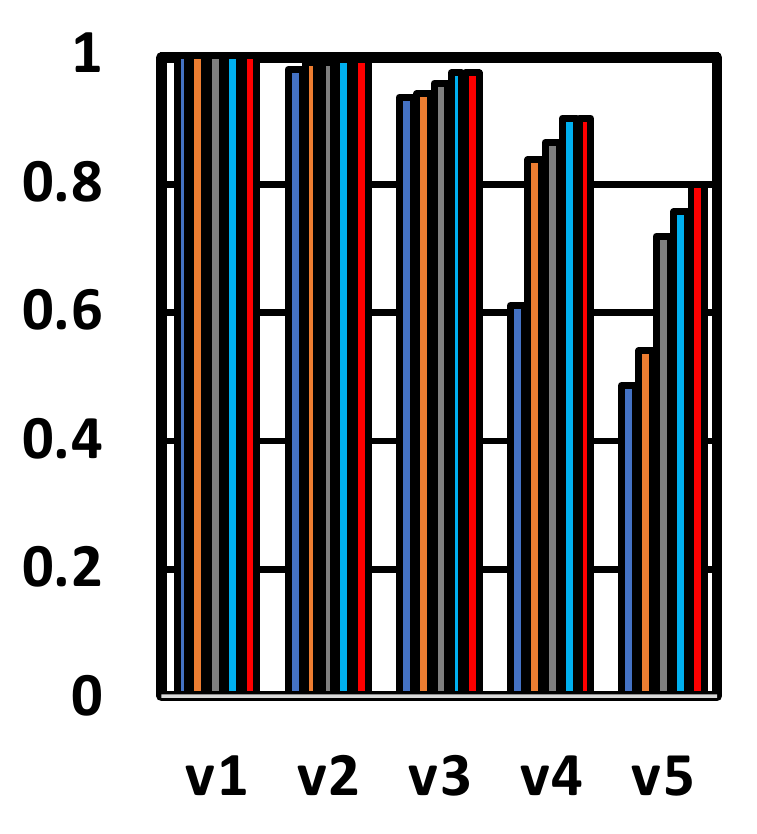}
    \caption{Model Depth 4}
  \end{subfigure}
  \begin{subfigure}{3cm}
    \centering\includegraphics[width=3cm]{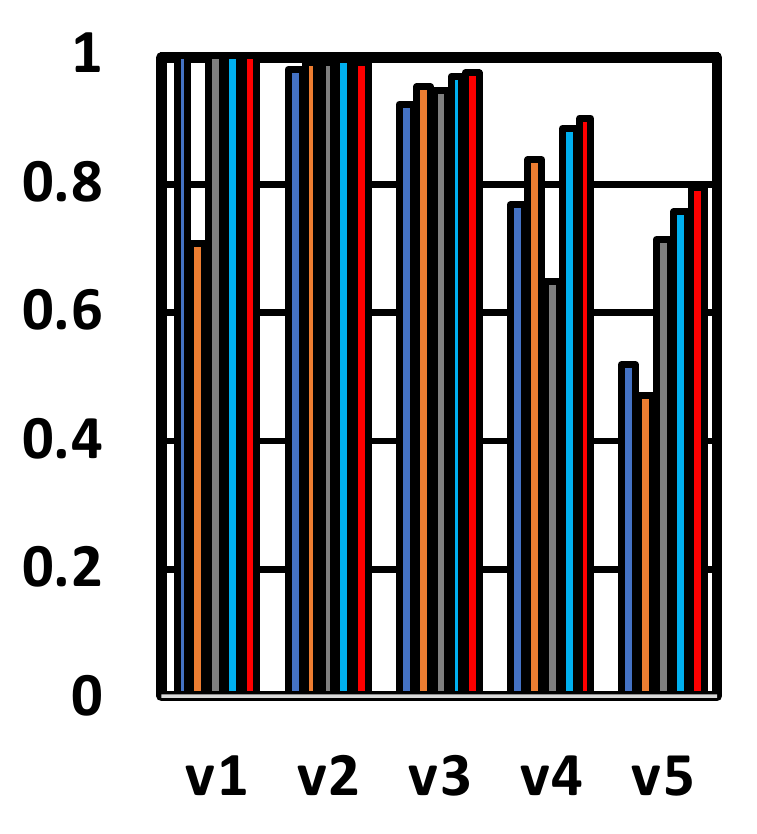}
    \caption{Model Depth 5}
  \end{subfigure}
  \caption{MLP generated Macro G-Mean on Gaussian Backbone: (a) Model Depth 1, (b) Model Depth 2, (c) Model Depth 3, (d) Model Depth 4, and (e) Model Depth 5. 
  Increasing the depth of the model up to depth 4 is beneficial, but the move to depth 5 deteriorates some of the classification performance.
  }
  \label{fig:GaussianBackboneMLPGMean}

\end{figure*}

The results suggest that combining overlap and imbalance in a moderately complex domain is detrimental to classification performance, but that increasing the number of hidden layers is beneficial, but only up to a certain depth. Indeed, while there is clear improvement from Figure \ref{fig:GaussianBackboneMLPGMean} (a) to (b) and then again (b) to (c), the improvement is minimal for (c) to (d) and the move from (d) to (e) causes more performance deterioration than improvement (see, for example the results for v1 and v4 in Figure \ref{fig:GaussianBackboneMLPGMean} (e)).

\section{The impact of Depth on the Class Imbalance Problem in image domains}
\label{sec:imageExps}

\noindent {\bf Description of the Experimental Setup.} The experiments described in the previous three sections were an attempt to characterize the role of neural networks' depth on the class imbalance problem affected by a variety of characteristics. We found that increasing the depth of a neural network is a useful way to improve classification results in class imbalance settings, but that it is not a panacea. Indeed as structural concept complexity, class overlap degree of imbalance and data scarcity increase, increasing the depth of the network is insufficient to improve classification and, as seen in the last set of experiments, may stop helping or even start deteriorating the performance.  
In this section, we seek to understand whether the results we have obtained on the artificial domains map to real world conditions or whether other factors that were not considered in our artificial data experiments are at play.

We consider two image domains to see how the conclusions drawn from the artificial domains' experiments translate to realistic domains and whether additional considerations need to be taken. For these experiments, two benchmark datasets are considered: Fashion-MNIST\footnote{\url{https://github.com/zalandoresearch/fashion-mnist}} and CIFAR10\footnote{\url{https://www.cs.toronto.edu/~kriz/cifar.html}}. 
Once again, we considered 5 balance levels (imbalance ratios) [level]: 0.5 (0) [b=5], 0.3 (3.33) [b=4], 0.15 (6.66) [b=3], 0.05 (20) [b=2], 0.025 (40) [b=1]. We created the minority classes of levels b=4..1 by randomly undersampling the second class (considered the minority class) of each binary domain. 




\smallskip
\noindent {\bf Domains.} Binary domains were selected using a combination of visual inspection of binary T-SNE plots and cross-validation experiments.
The aim of the selection was to identify five binary domains with increasing levels of complexity: starting from an easy domain where points appear mostly linearly separable, to a moderate domain characterized by the structural concept complexity and overlapping phenomena, to more extreme and difficult scenarios that suffer from both issues simultaneously. 

In order to select the domains, we sampled 500 images from each class and exhaustively generated plots for all pairwise combinations of classes. After visually pre-selecting the most relevant domains in each category, we performed experiments to confirm the complexity of the selected scenarios and rank them accordingly. In particular, we adopted the average G-Mean performance achieved using 2x10-fold stratified cross-validation\footnote{2x10-fold stratified cross-validation is used to ensure stable means and standard deviations in the results.} on the binary settings with different model architectures (from 1 up to 5 convolutional layers) using balanced data. The selected domains ordered by increasing level of complexity are shown in Figure \ref{fig:mnistFashionDataPlots} and \ref{fig:cifar10DataPlots}.

\begin{figure*}[t]
  \centering
    \centering\includegraphics[scale=0.255]{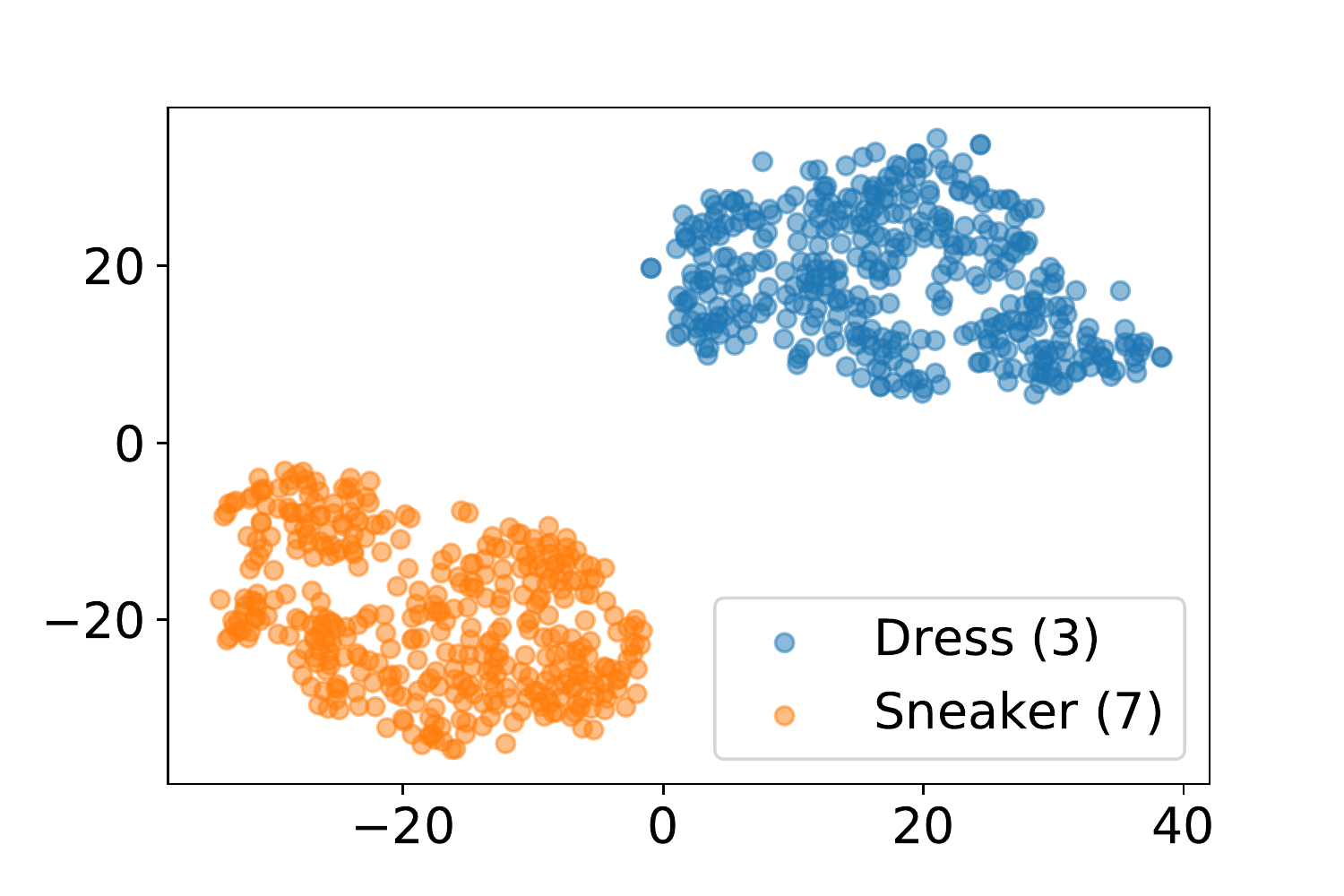}
    \hspace{-0.6cm}
    \centering\includegraphics[scale=0.255]{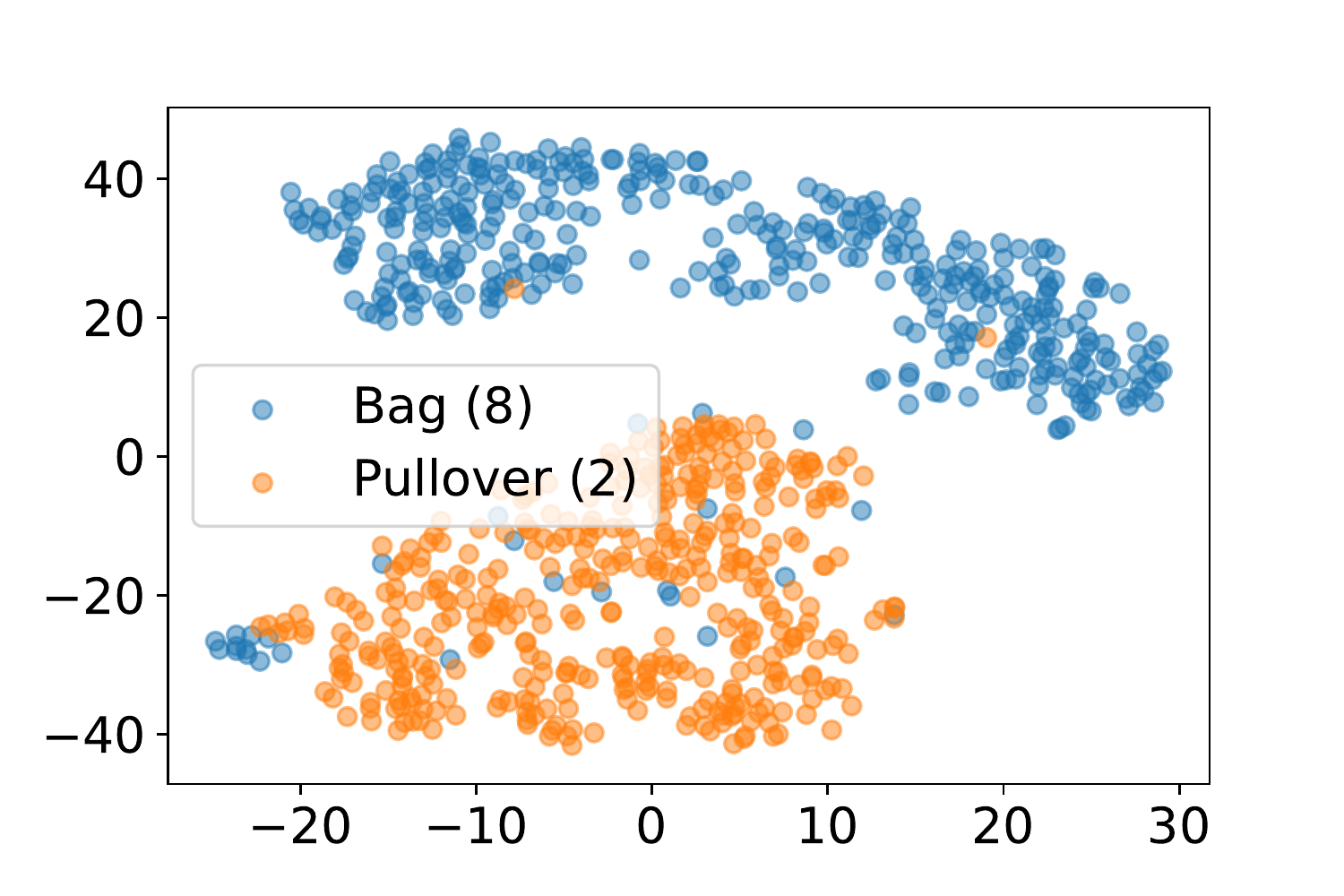}
    \hspace{-0.6cm}
    \centering\includegraphics[scale=0.255]{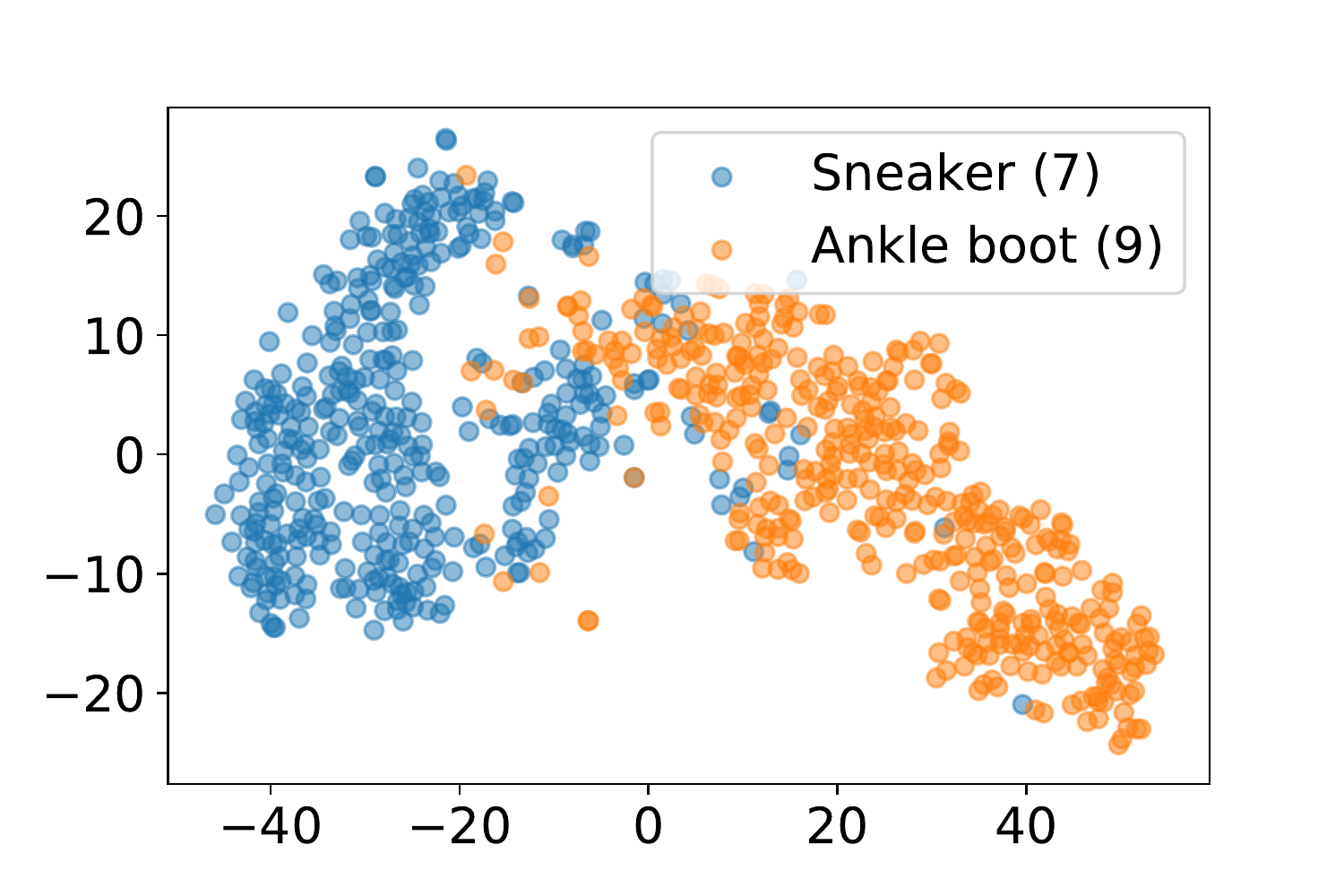}
    \hspace{-0.6cm}
    \centering\includegraphics[scale=0.255]{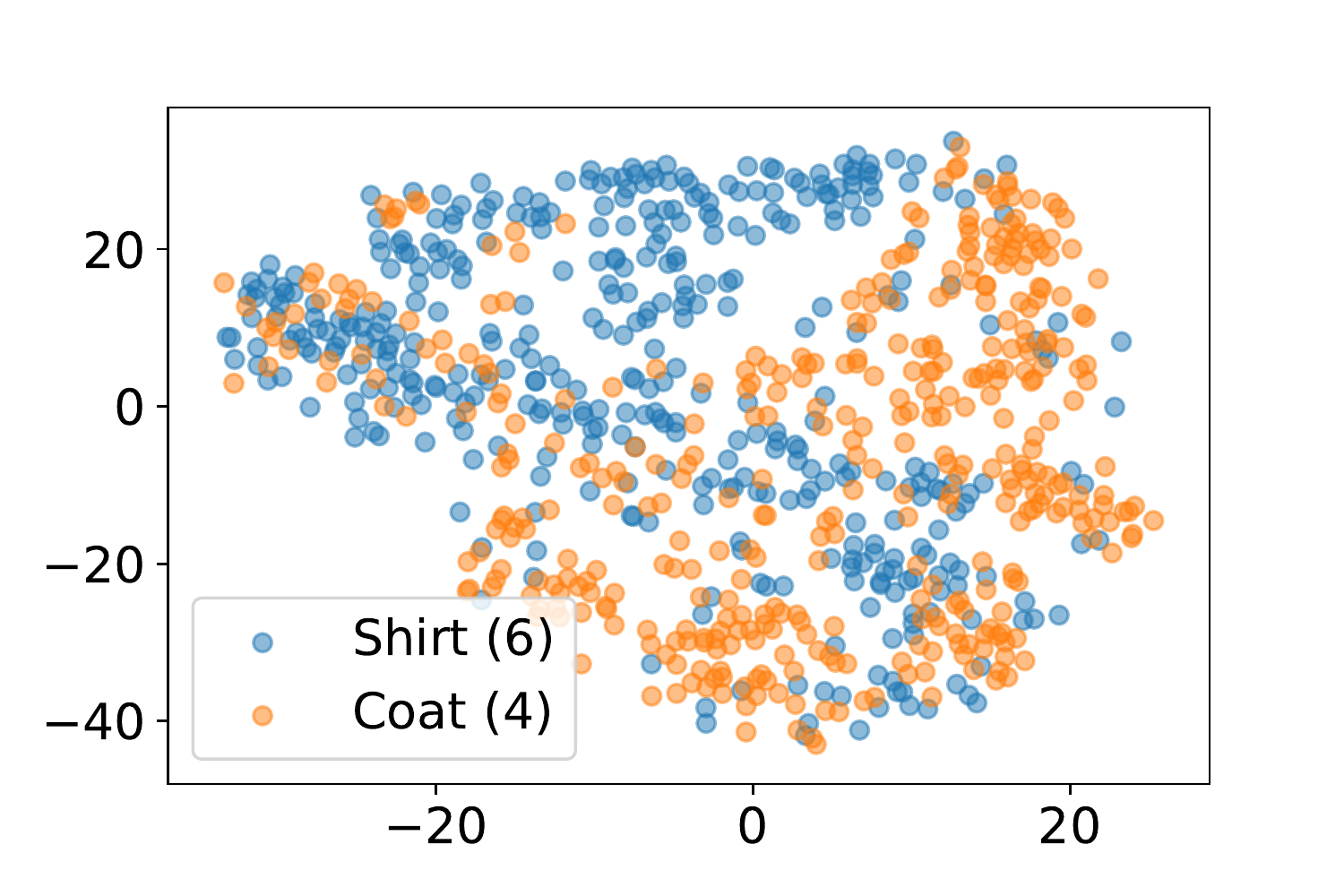}
    \hspace{-0.6cm}
    \centering\includegraphics[scale=0.255]{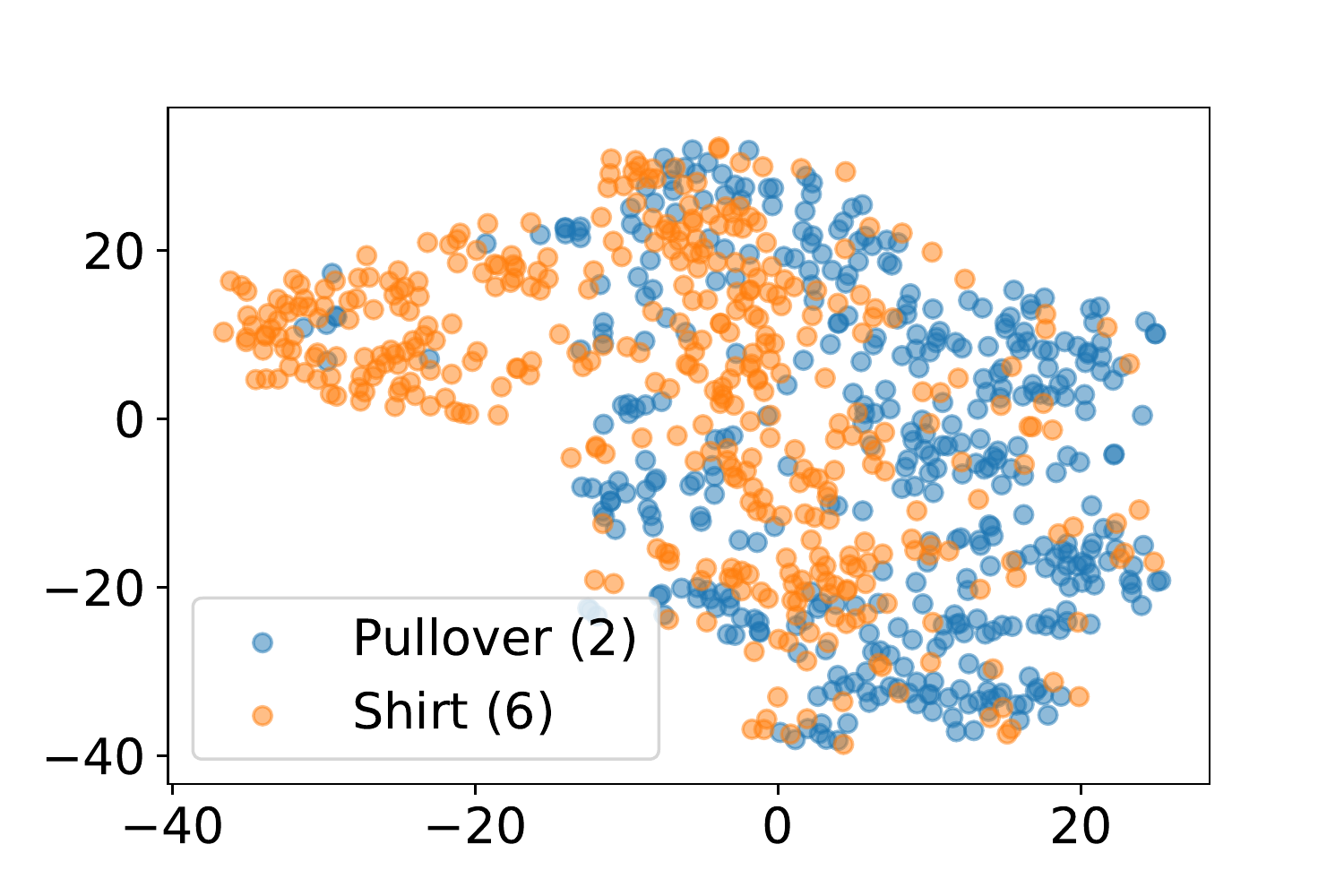}
  \caption{T-SNE Plots of Binary MNIST Fashion datasets sorted from the least to the most complex.}
  \label{fig:mnistFashionDataPlots}
\end{figure*}

 \begin{figure*}[t]
     \centering\includegraphics[scale=0.24]{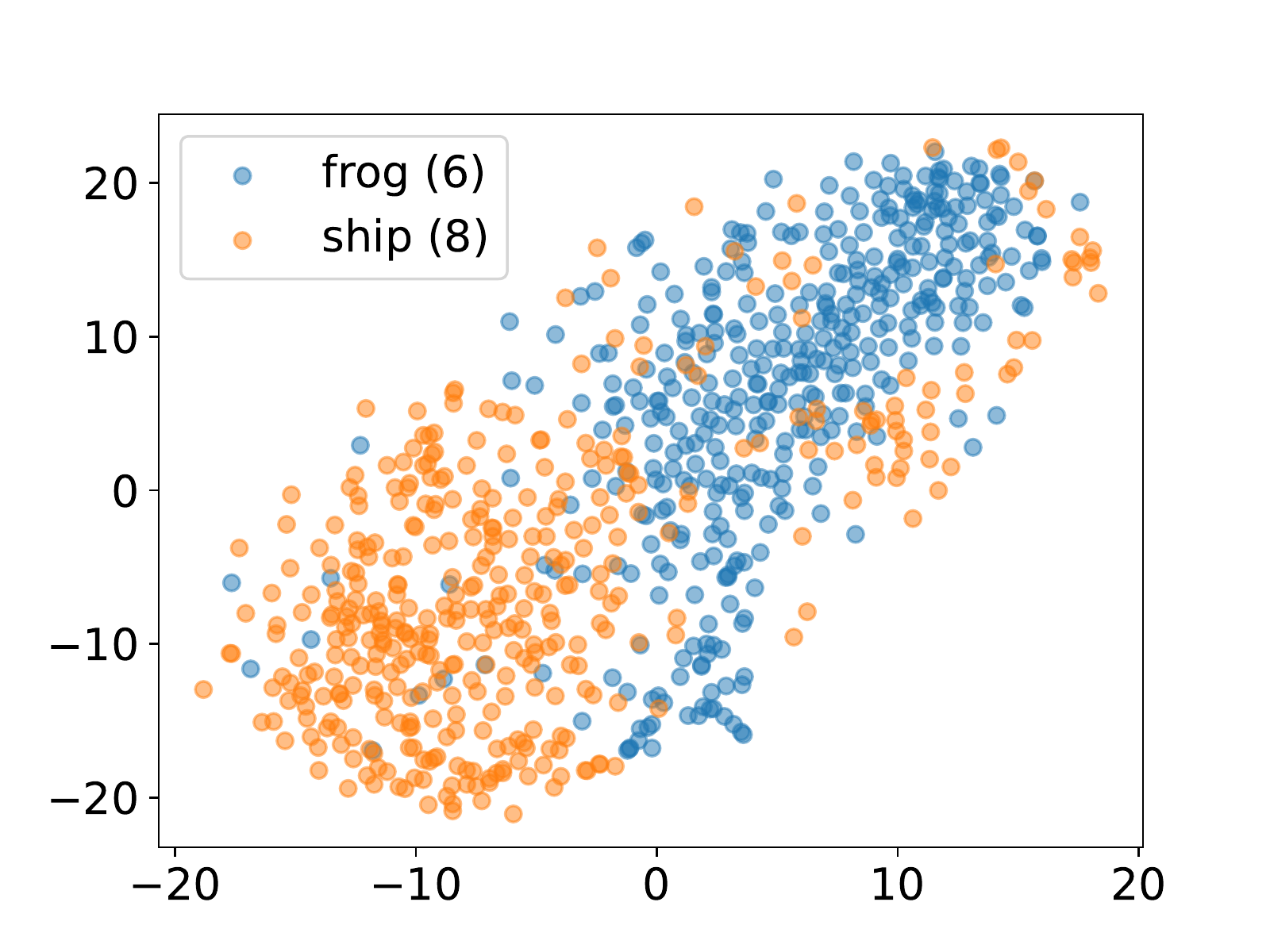}
     \hspace{-0.63cm}
     \centering\includegraphics[scale=0.24]{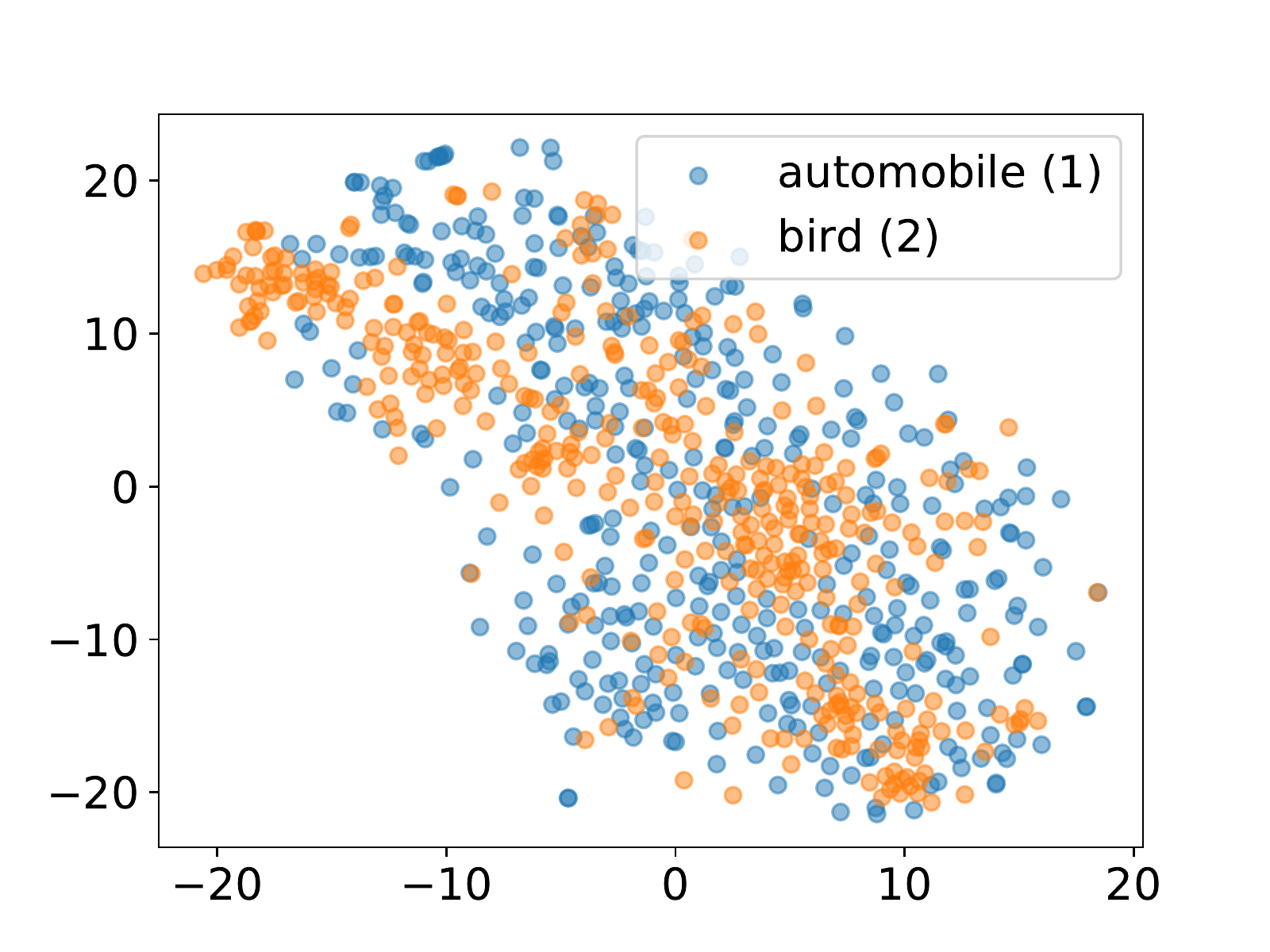}
     \hspace{-0.63cm}
     \centering\includegraphics[scale=0.24]{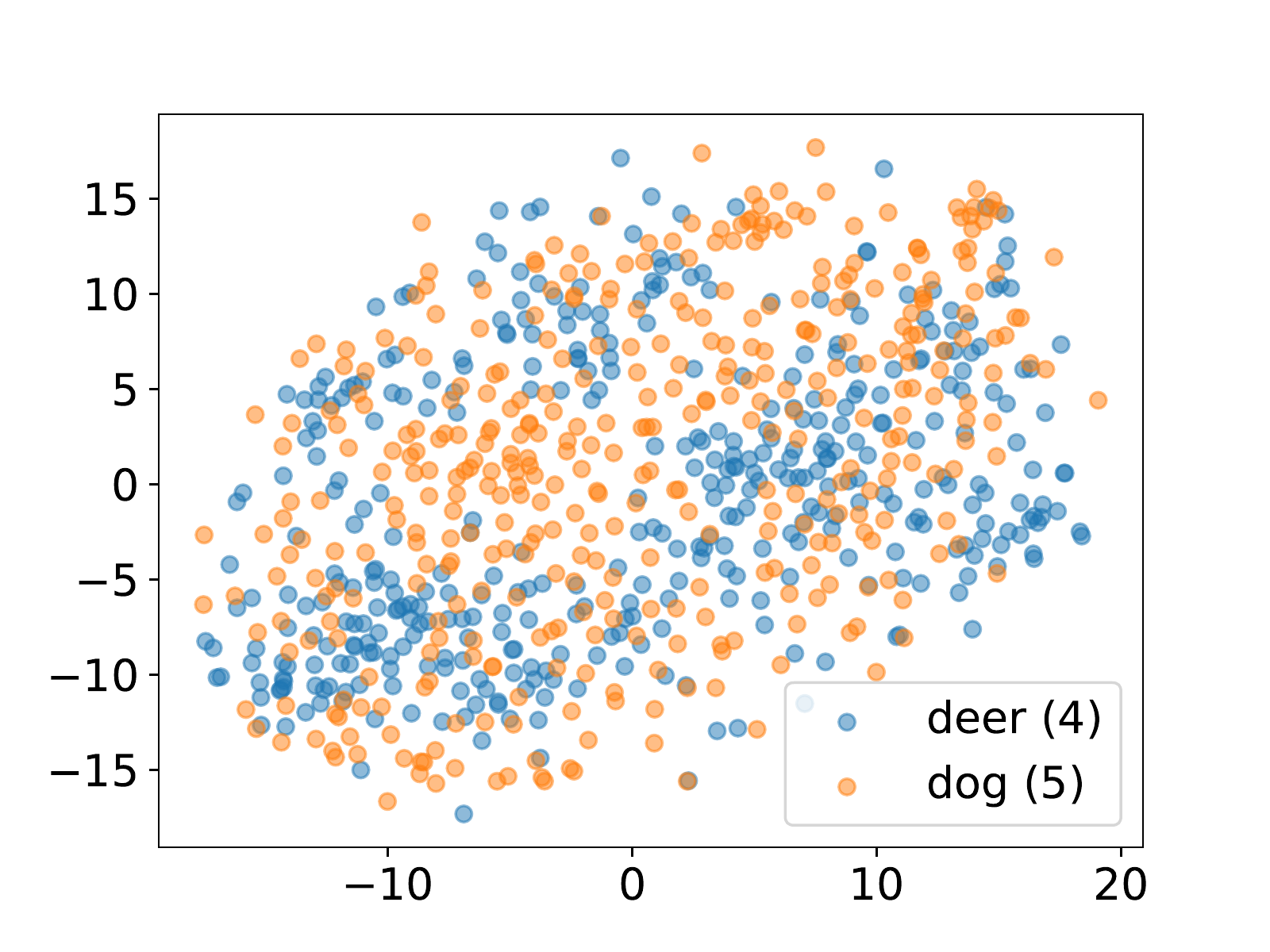}
     \hspace{-0.63cm}
     \centering\includegraphics[scale=0.24]{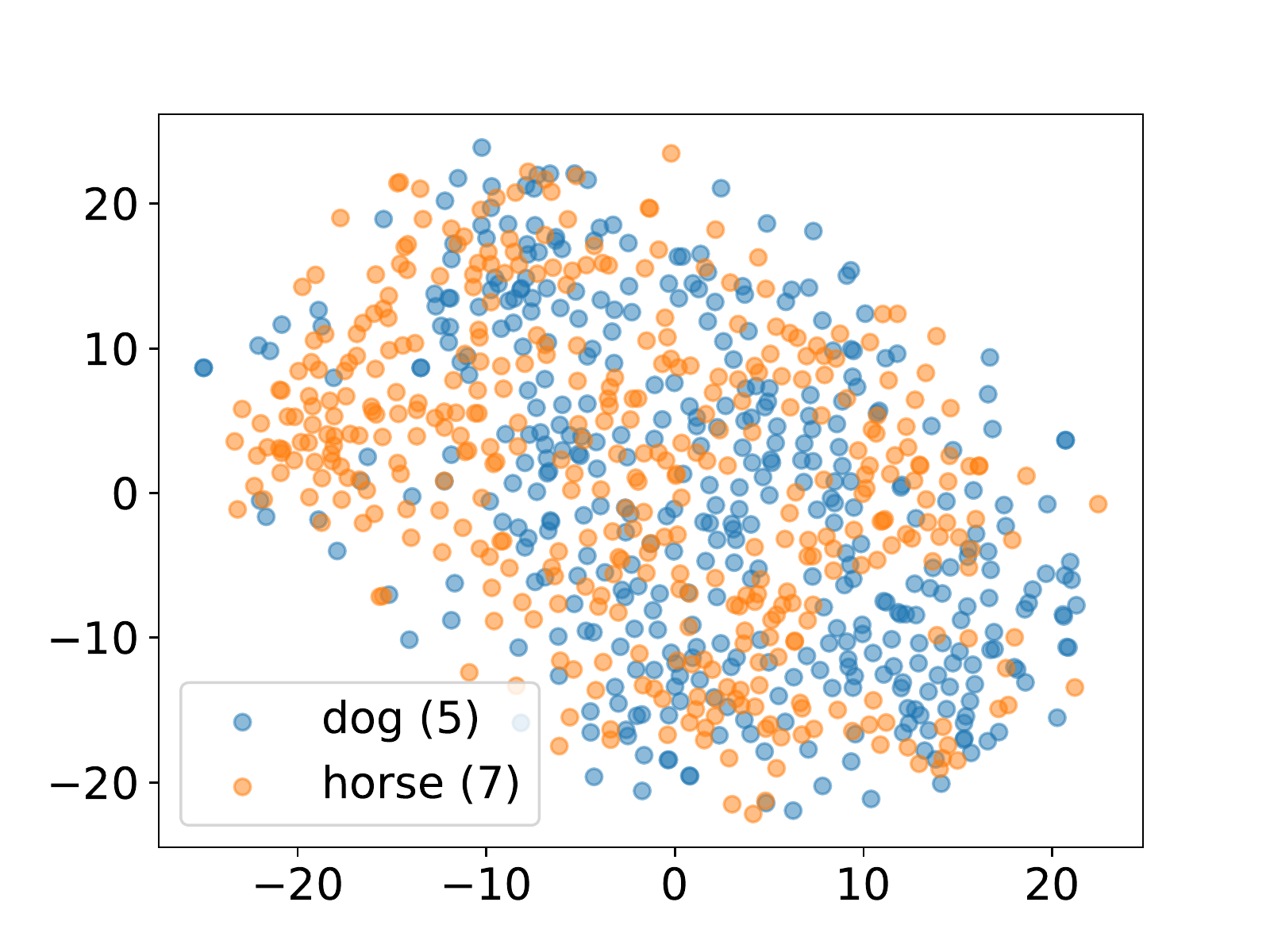}
     \hspace{-0.63cm}
     \centering\includegraphics[scale=0.24]{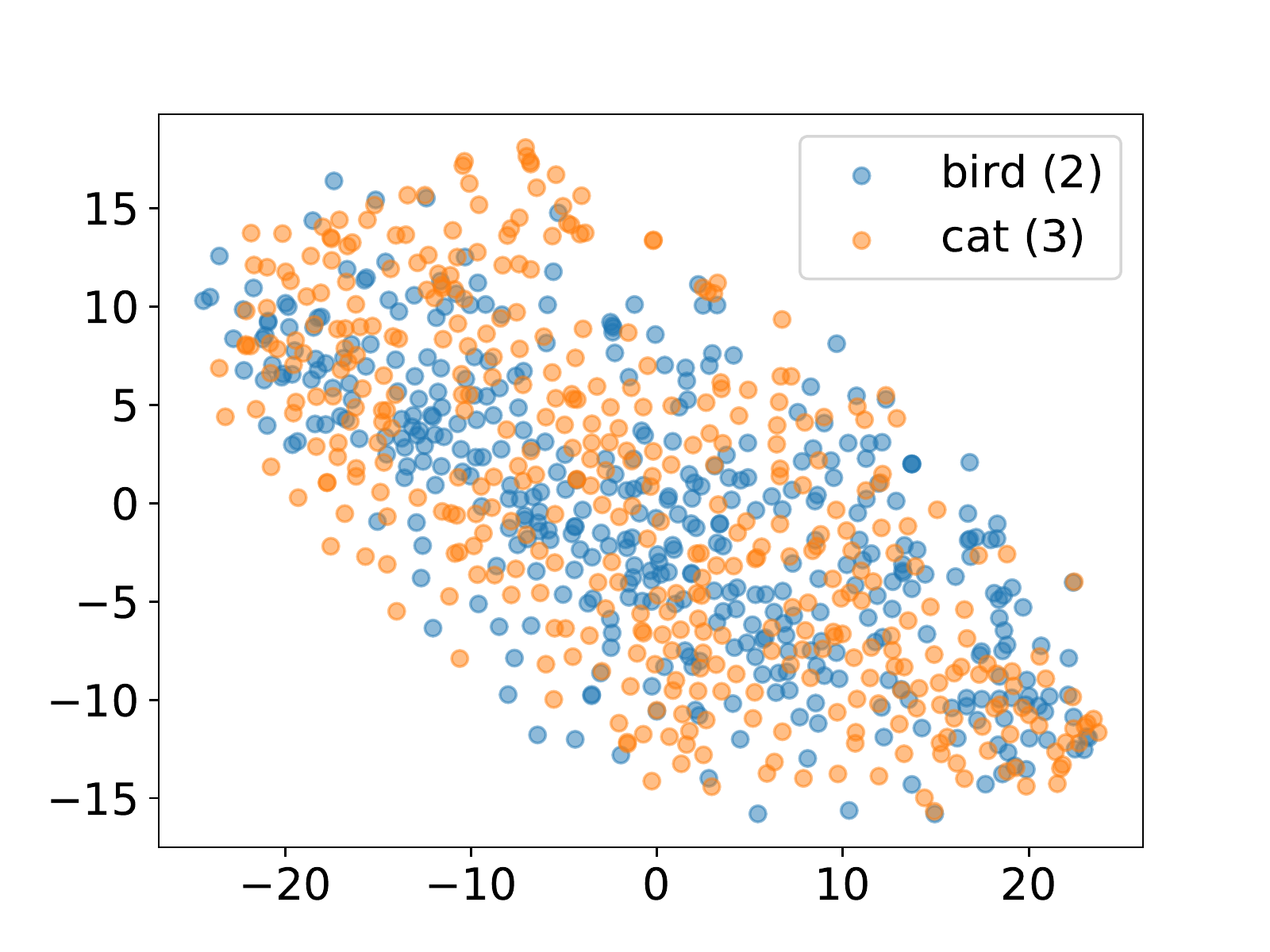}
   \caption{T-SNE Plots of Binary CIFAR10 datasets sorted from the least to the most complex.}
   \label{fig:cifar10DataPlots}
 \end{figure*}
 
 \smallskip
\noindent {\bf Models and their parameters.} The model architecture considered in this set ofexperiments is a Convolutional Network with an increasing number of convolutional layers (filters): 1 (8), 2 (8-16), 3 (8-16-32), 4 (8-16-32-64), 5 (8-16-32-64-64). Two dense layers are featured at the end of each model architecture.

\smallskip
\noindent {\bf Results.} The results obtained on MNIST Fashion can be found in Figure \ref{fig:mnistFashionDataPlots}, while those obtained on CIFAR-10 can be found in Figure \ref{fig:cifar10DataPlots}. Both figures are organized in the same fashion as Figures \ref{fig:9} and \ref{fig:10} in the artificial domains. In other words, each subplot(a) to (e) represent the results obtained with increasing depth going from 1 to 5, respectively; within each subplot, clusters of bars show a different degree of difficulty which can be visualized in the plots of Figures \ref{fig:mnistFashionGMean} and \ref{fig:cifar10GMean} for MNIST Fashion and CIFAR-10, respectively; and within each cluster of bars, from left to right, the height of the bar represents the G-Mean performance obtained from the least balanced data set (2.5\%minority) to the perfectly balanced data set  (50\%minority). While the format of the results follows that of the artificial structural concept complexity results, it is important to note, that, in fact, overlap is also considered. This is because, in the real-world image domains considered, both the notions of overlap and structural concept complexity are mixed as they were in the experiments of section IV. This is evident in Figures \ref{fig:mnistFashionDataPlots} and \ref{fig:cifar10DataPlots} where subclusters can be seen in some of the overlapping regions and overlap due to high variance and/or closely located subcluster means is simultaneously present. The presence of both phenomena is particularly visible in Figure \ref{fig:mnistFashionDataPlots} (d) and (e) and Figure \ref{fig:cifar10DataPlots} (b)-(e). In this respect, this experiment is best modeled by the experiment on artificial data from Section IV.

\begin{figure*}[t]
  \centering
  \begin{subfigure}{3cm}
    \centering\includegraphics[width=3cm]{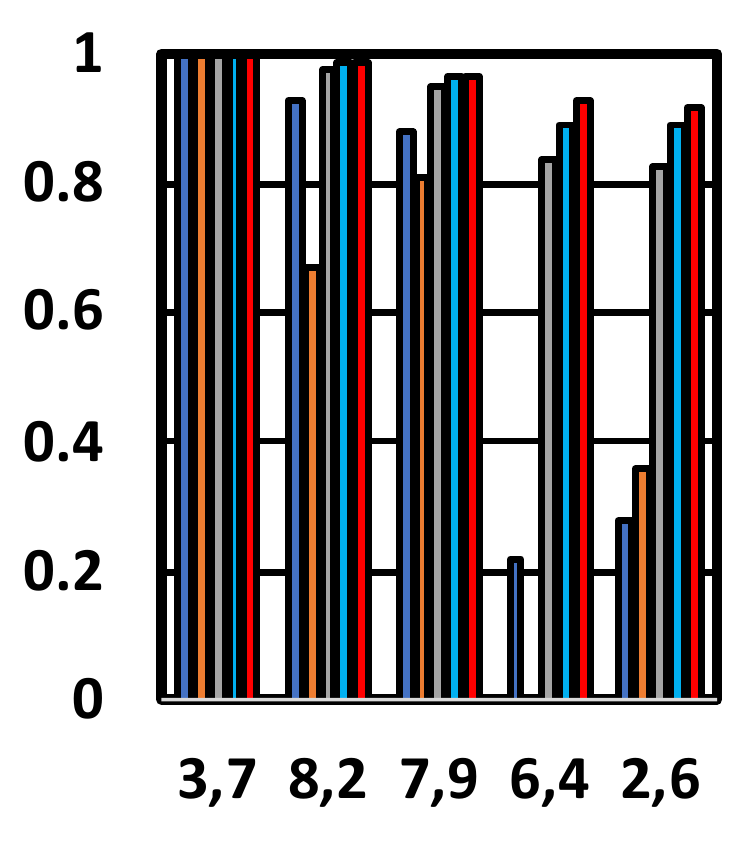}
    \caption{Model Depth 1}
  \end{subfigure}
  \begin{subfigure}{3cm}
    \centering\includegraphics[width=3cm]{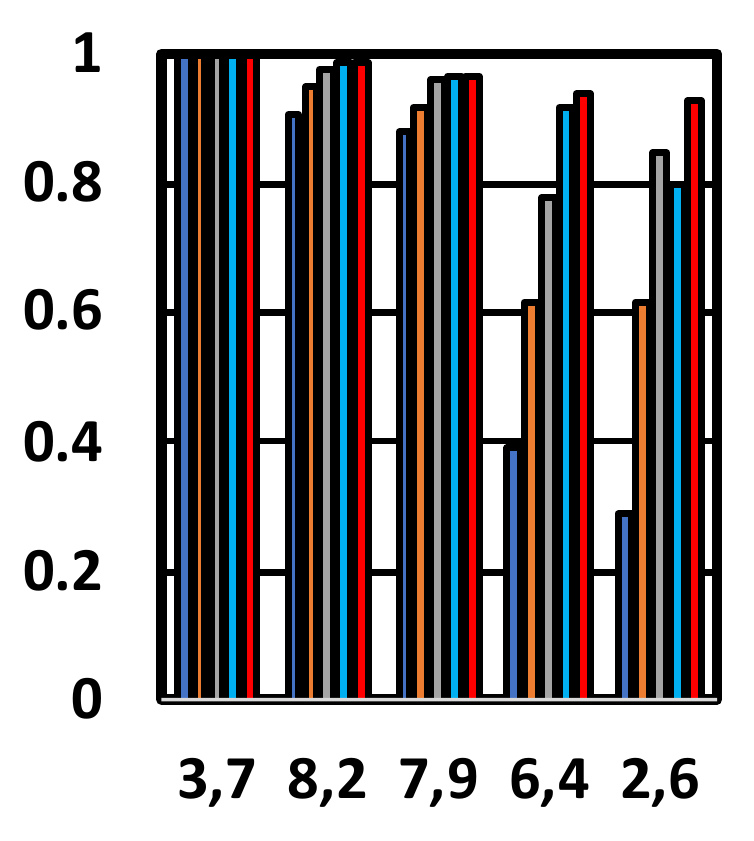}
    \caption{Model Depth 2}
  \end{subfigure}
  \begin{subfigure}{3cm}
    \centering\includegraphics[width=3cm]{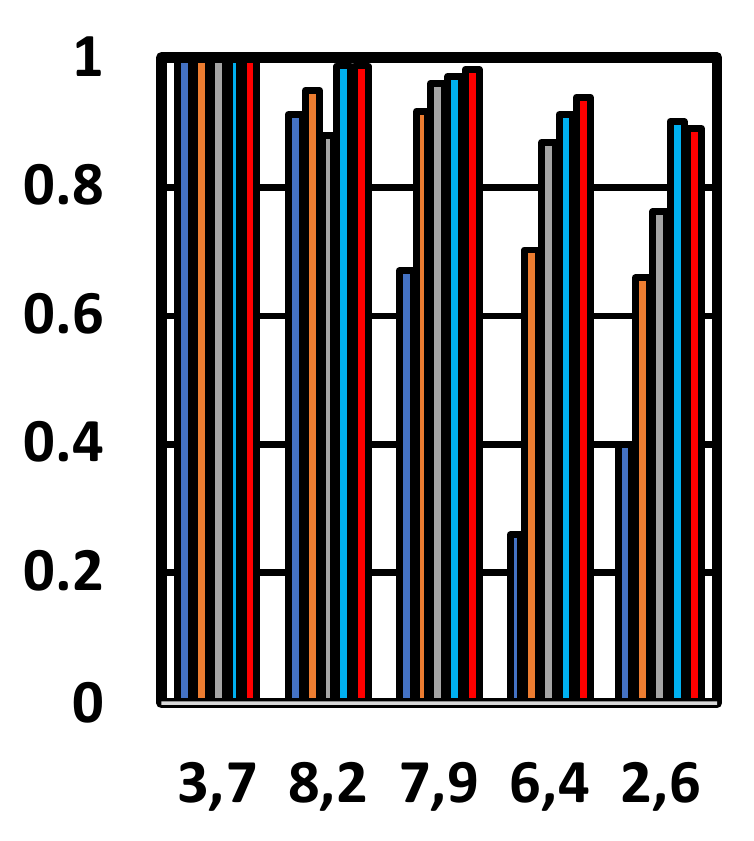}
    \caption{Model Depth 3}
  \end{subfigure}
  \begin{subfigure}{3cm}
    \centering\includegraphics[width=3cm]{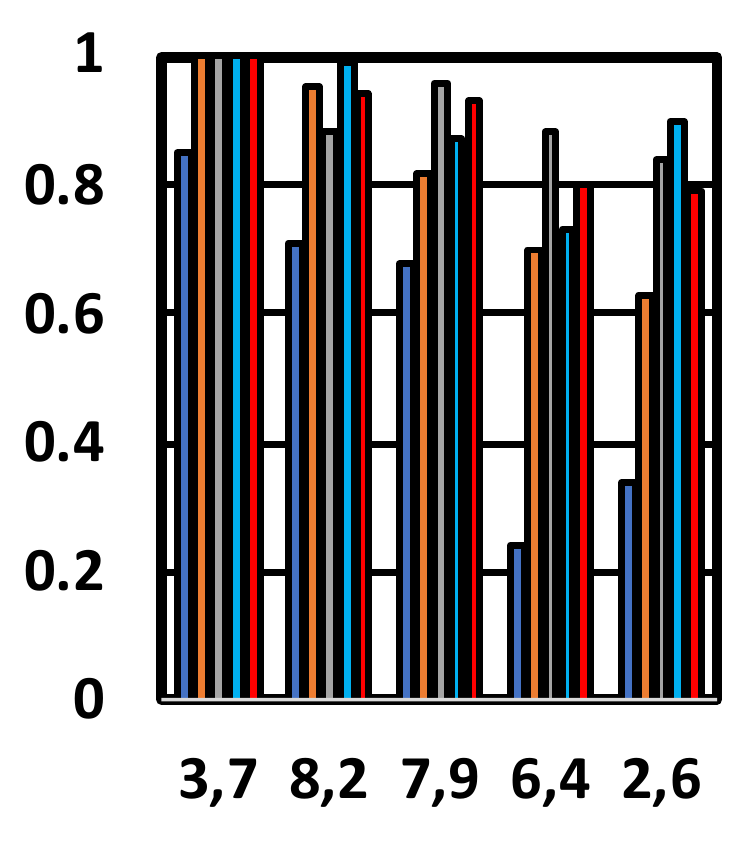}
    \caption{Model Depth 4}
  \end{subfigure}
  \begin{subfigure}{3cm}
    \centering\includegraphics[width=3cm]{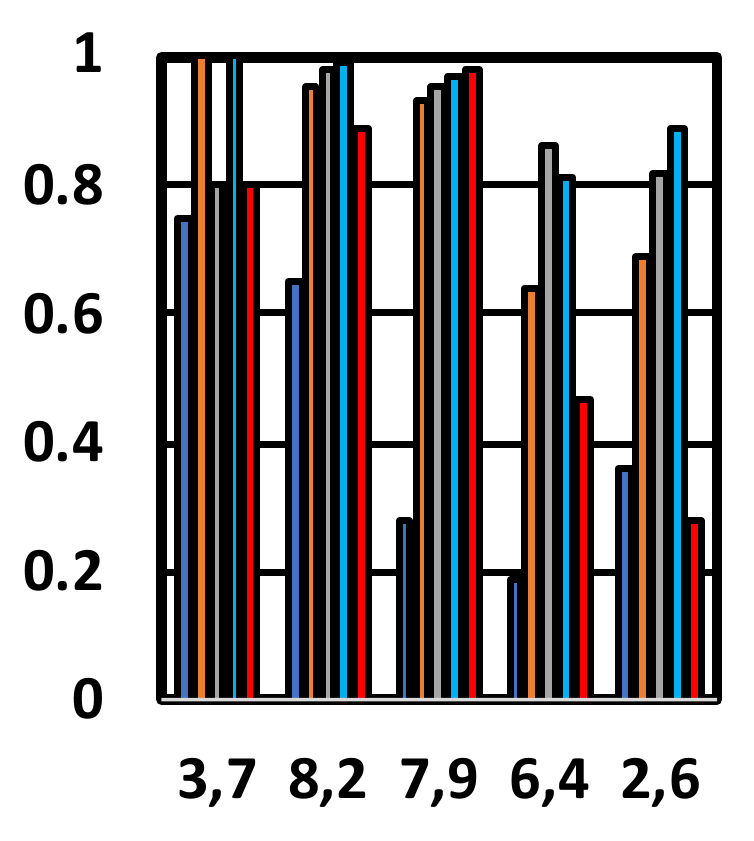}
    \caption{Model Depth 5}
  \end{subfigure}
  \caption{G-Mean on MNIST Fashion: (a) Model Depth 1, (b) Model Depth 2, (c) Model Depth 3, (d) Model Depth 4, and (e) Model Depth 5.}
  \label{fig:mnistFashionGMean}
\end{figure*}

\begin{figure*}[t]
  \centering
  \begin{subfigure}{3cm}
    \centering\includegraphics[width=3cm]{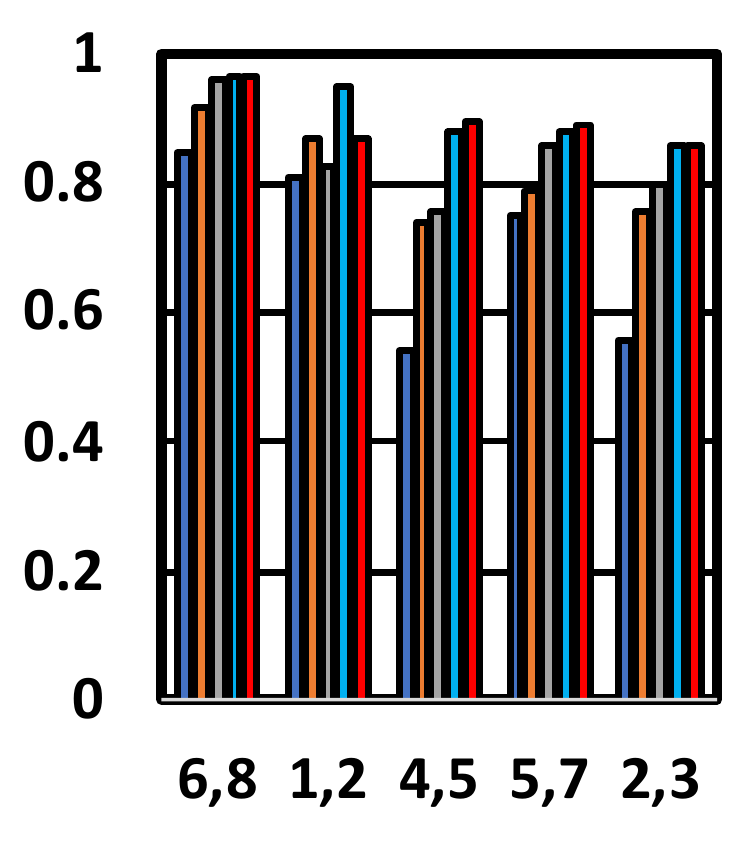}
    \caption{Model Depth 1}
  \end{subfigure}
  \begin{subfigure}{3cm}
    \centering\includegraphics[width=3cm]{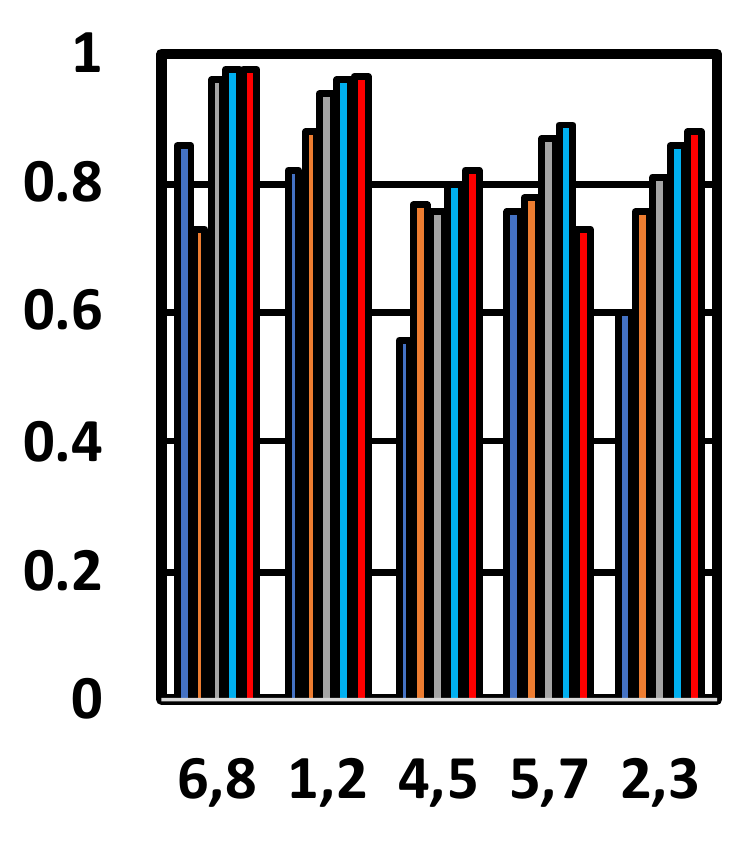}
    \caption{Model Depth 2}
  \end{subfigure}
  \begin{subfigure}{3cm}
    \centering\includegraphics[width=3cm]{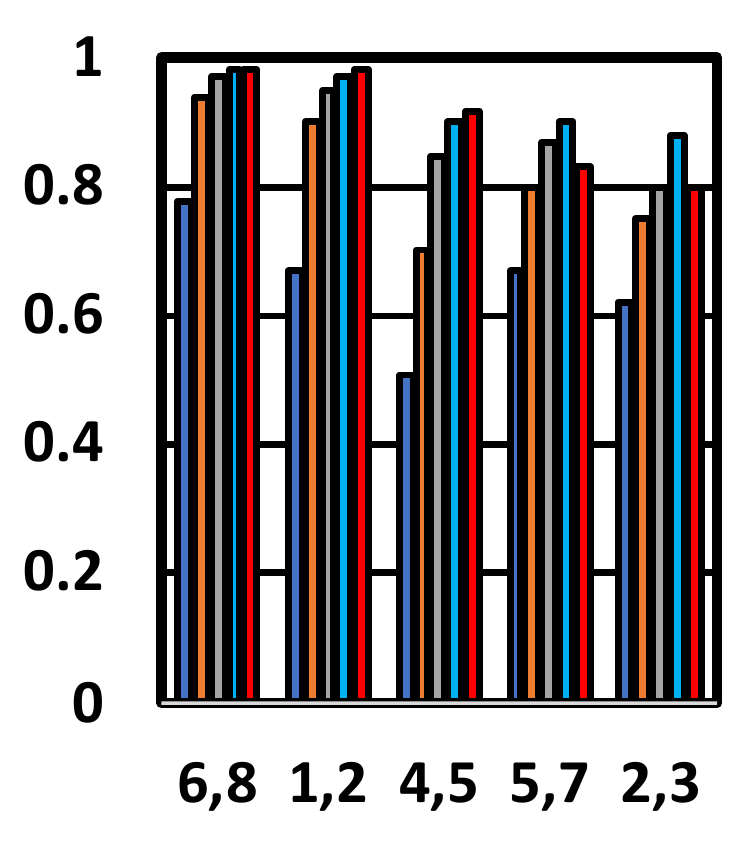}
    \caption{Model Depth 3}
  \end{subfigure}
  \begin{subfigure}{3cm}
    \centering\includegraphics[width=3cm]{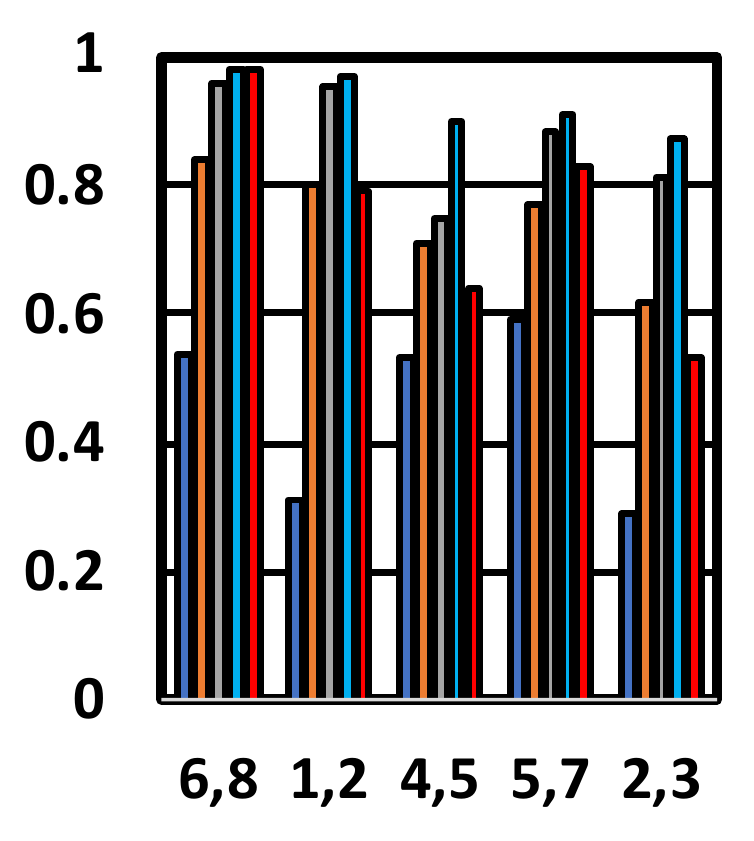}
    \caption{Model Depth 4}
  \end{subfigure}
  \begin{subfigure}{3cm}
    \centering\includegraphics[width=3cm]{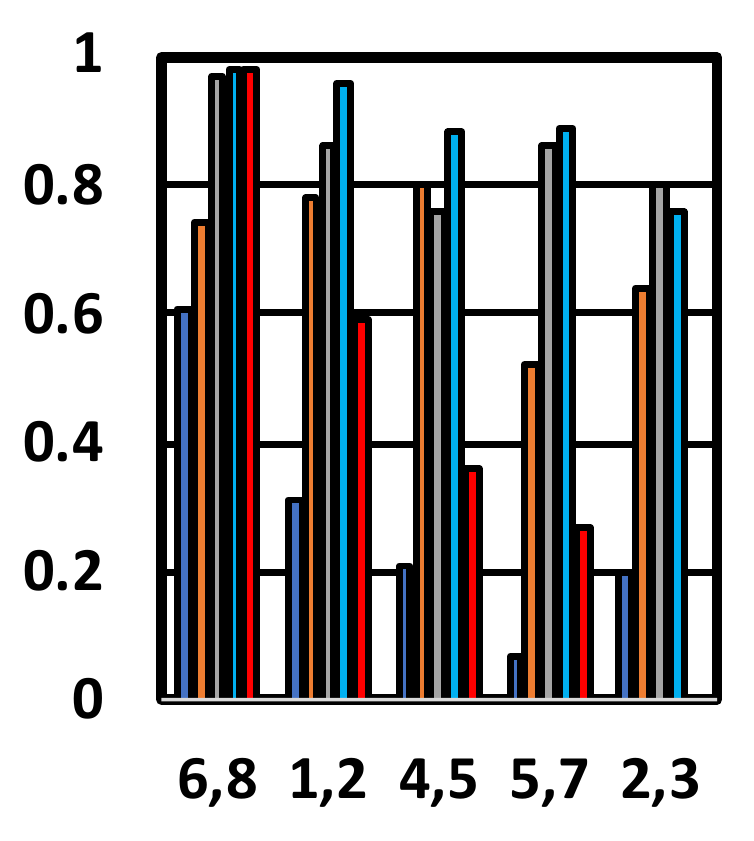}
    \caption{Model Depth 5}
  \end{subfigure}
  \caption{G-Mean on CIFAR10: (a) Model Depth 1, (b) Model Depth 2, (c) Model Depth 3, (d) Model Depth 4, and (e) Model Depth 5.}
  \label{fig:cifar10GMean}
\end{figure*}

Our first major observation, which once again serves as a prelude to our main question,  is that, like in the case of our concept complexity experiments on artificial domains, as the degrees of domain complexity and class imbalances increase, the performance of the classifiers decreases. For example, in the MNIST Fashion data, in Figure \ref{fig:mnistFashionGMean} (a), for concept complexities c1 (3, 7) or c2 (8, 2), there is virtually no difference between the performance obtained on the completely balanced or most imbalanced data set (except for c2 (8, 2) at balance level b=2 (Minority class 5\%), but this results seems to be a random aberration). For concept complexity c4 (6, 4) and c5 (2, 6), however, there is a difference of around 70\% in G-Mean between balance level b=1 and b=5. Because of the higher degree of complexity of all the domains in the CIFAR-10 experiments as observed in the graphs of Figure \ref{fig:cifar10DataPlots}, the difference between simpler and more complex domains is not as well marked as it is in the MNIST Fashion datasets as can be seen in Figure~\ref{fig:cifar10GMean}. Nonetheless, we can observe a clear difference between the domain corresponding to the simplest complexity level c1 (6, 8) and all the others. For the simplest domain, c1 (6, 8), the difference in G-Mean between balance levels b=1 and b= 5 for one hidden layer is 12\% whereas for all the other more complex domains c2 (1, 2) - c5 (2, 3) (except c3 (4, 5), for some reason), they hover around 30\%. 

Our second major observation, which relates to the main question in the paper, concerns the number of layers used in our networks. The pattern, in this case, is less clear than it was in the artificial domains, though it ressembles, somewhat, the results obtained in Section IV. In the most extreme cases such as in MNIST Fashion's concept complexities c4 (6, 4) or c5 (2, 6) and degree of balance b=1, but even in some other cases such as concept complexity c2 (8, 2) and balance level b=2, we notice a pattern where the addition of one or two layers helps improve the G-Mean. Adding more than 2 layers, however, hurts the performance. In other MNIST Fashion cases as well as in the case of the CIFAR-10 data, the results are closer to what we observe in the artificial domains concerned with overlap alone: additional layers do not help. In fact, as we add more layers, there are many cases where the performance decreases significantly. In the rare cases in both MNIST Fashion and the CIFAR-10 datasets where the addition of  layer helps, it does so only slightly. 

The difference between our artificial and realistic domain observations, we believe, is related to the fact that while increasing layers may help learn more complex boundaries between classes, it also causes the systems to overfit the data. Because of the simplicity of the artificial domains, the overfitting issue may not affect the performance of the deep networks on these domains
 while it dominates the performance of the deep networks in the real image domains. In these domains, any benefit brought upon by deeper layers is overshadowed by the amount of overfitting that it also causes.  In fact, this can be seen as a classical case of the bias and variance dilemma where, the decrease in bias obtained through deeper networks, is also responsible for an increase in variance. Further investigation, however, is necessary to establish a formal explanation of the observed disparity.


\smallskip
\noindent {\bf CNN and MLP Layer Embedding Analysis.} In this section, we examine the representations learned by the convolutional (CNN) and dense (MLP) layers of the network used on the real world domains in order to understand how they are impacted by concept complexity, class overlap and imbalance. To achieve this, we focus on 
a `hard' classification problem ( concept complexity c5, or class 2 versus class 6) 
from the MNIST Fashion dataset. 

To undertake this analysis, for each dataset, we train CNNs with 1, 3 and 5 convolution blocks (as discussed in Section \ref{sec:imageExps}) with a balanced class setup (b=5)  and an extreme imbalanced class setup (2.5\% imbalance level, or b=1). In order to provide clear picture of the efficacy of the learned model, we extract representations from an independent balanced test set from each model for visualization. The representations are recorded after the final convolutional block (CNN) and after the final fully connected layer (MLP). We utilize T-SNE \cite{van2008visualizing} as the dimension reduction technique to plot the data and learned representation in 2 dimensions. 

\begin{figure*}[t]
  \centering
    \centering\includegraphics[width=4cm]{figures/mf2versus6.pdf}
  
    \centering\includegraphics[width=4cm]{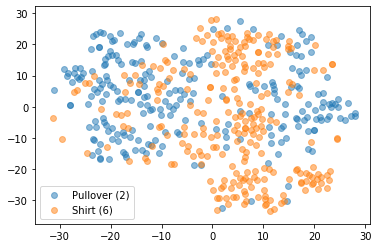}
    \centering\includegraphics[width=4cm]{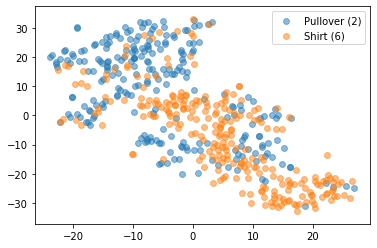}
    \centering\includegraphics[width=4cm]{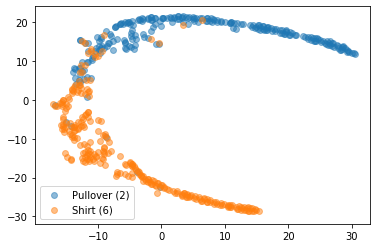}
    \centering\includegraphics[width=4cm]{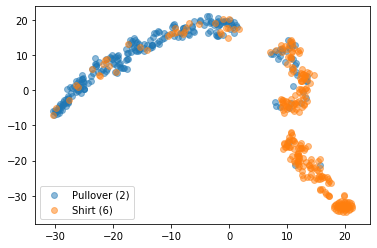}
    
    \centering\includegraphics[width=4cm]{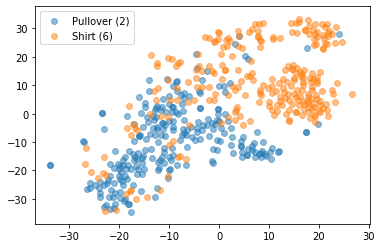}
    \centering\includegraphics[width=4cm]{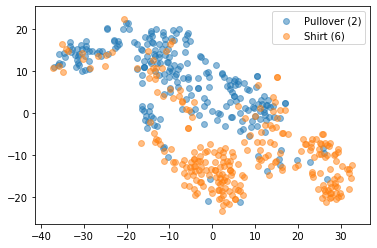}
    \centering\includegraphics[width=4cm]{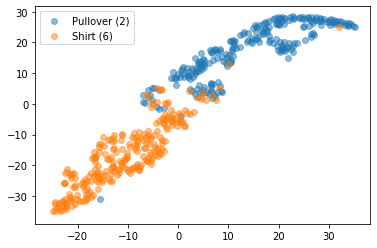}
    \centering\includegraphics[width=4cm]{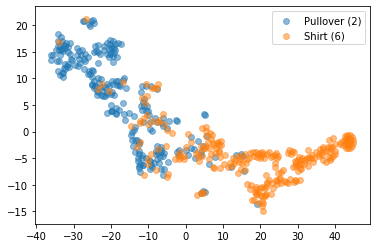}
    
    \centering\includegraphics[width=4cm]{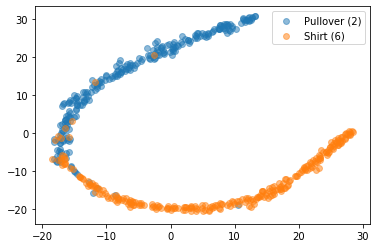}
    \centering\includegraphics[width=4cm]{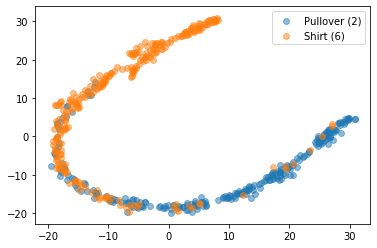}
    \centering\includegraphics[width=4cm]{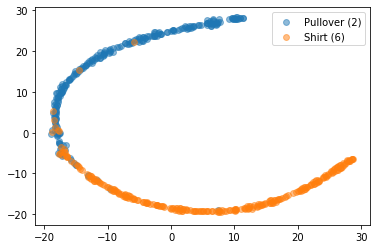}
    \centering\includegraphics[width=4cm]{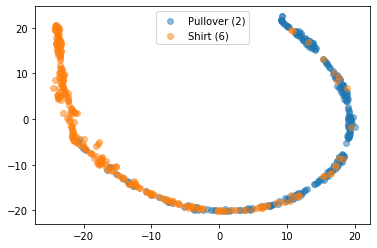}

  \caption{These plots illustrate the representation learned on the MNIST Fashion for classes 2 versus 6. The plot on the top row shows the T-SNE embedding of the original data. This reveals that the classification problem is complex due to the presence of class overlap and sub-concepts. In the subsequent rows, the first and second columns show the T-SNE plots for the representations learned after the last CNN block on the balanced (first column) and imbalanced (second column) data. The third and fourth columns show the T-SNE plots for the representations learned after the last fully connected layers for balanced (third column) and imbalanced (fourth column). Row-wise, the plot represent the results of CNNs with 1, 3, and 5 CNN blocks. These plots illustrate that adding more CNN blocks on the balanced data improves separability of the classes. However, whilst 3 blocks in the imbalanced case appears to slightly improve separability, 5 blocks exacerbates the class overlap.}
  \label{fig:cnnMnistFashion2_6Embeddings}
\end{figure*}

The results of this analysis are shown in Figure~\ref{fig:cnnMnistFashion2_6Embeddings} where visual inspection of the plots illustrates that whilst increasing the number of convolutional blocks can improve the separability and compactness of the representations learned on balanced, complex and non-complex data, it negatively impacts the representation learned when training on imbalanced data. It is particularly noteworthy that the representations learned with five convolutions blocks for the easy imbalanced classification problem (class 3 versus 7) causes class overlap where previously there was none\footnote{A detailed figure showing what happens in a simple domain is available upon request.}.

\section{Conclusions and Future Work}
\noindent {\bf Summary of the Study.} The purpose of this work was to investigate the response of neural networks of increasing depth to well-known harmful data challenges such as the combination of class imbalances, structural concept complexity, data scarcity, and class overlap. The study was conducted on a large number of artificial domains displaying combinations of different challenges expressed at various degrees as well as on two  image datasets very commonly used in the deep learning community. The architectures investigated were feed-forward multi-layer perceptrons (MLP) 
in the artificial domains as well as combinations of both MLP and convolutional neural networks in the real-world domains. The combination of convolutional and dense (MLP) layers was necessary in the real-world domains since representational learning as well as classification boundary learning had to be conducted simultaneously. Conversely, the artificial domains were designed so as not to require substantial representational learning so that the study could focus on the class boundary learning ability of the MLP type of deep architectures.

\smallskip
\noindent {\bf Discussion of the Results.} Although data challenges such as class imbalances, long tail distributions, data scarcity and so on have previously been recognized in the deep learning community \cite{b3}, the focus there has been primarily on  developing methods to improve performance, through resampling, cost adjustment and new loss functions, rather than understanding how and why deep models are impacted. This is the first systematic study of deep learning architectures' ability to learn complex class boundaries under non-optimal settings.
Our results show that like their classical counterparts, deep learning systems suffer from class imbalances, structural concept complexity, data scarcity, and class overlaps. This is not a complete surprise, but our study showed the interaction between these different challenges in great detail thus shedding lights on certain details such as the fact that structural concept complexity is more challenging to deep learning architectures than relatively large overlaps. More revealing, however, was  the interaction between the different data challenges and the depth of the learning systems. While structural concept complexity and class imbalances are quite detrimental to the performance of deep architectures, especially in cases of data scarcity, the addition of extra layers clearly helped overcome some of that difficulty. Conversely, the addition of extra layers in overlapping domains was practically ineffective, although not detrimental. When both effects were combined, depth helped at first but seemed to cause harm as it increased too much. This effect was shown slightly in a family of artificial domains that mixed the two characteristics, but it was displayed  very clearly 
in the study on real image domains. There, it was found that as the domain complexity (of both structural concept complexity and overlap) increased, adding extra layers was frequently detrimental to the performance of the system, except in some cases where one or two extra layers (but not five!) could help a bit. This is an interesting results which, we believe, illustrates the tradeoff between bias and variance in the deep learning setting: while adding layers helps the networks capture the complexity of the boundaries, thus reducing the bias term, the effect of that addition is so detrimental to the variance term that any gains made on the bias term are cancelled out by the variance term. This is not necessarily the case in all real-world domains, however, as suggested by \cite{b4} who show that in the context of Facial action recognition, the use of very deep networks appears beneficial. This suggests that either those domains present characteristics quite different from the image domains we used and/or that the authors' use of regularisers may helped improve the performance of the deep architectures.

\smallskip
\noindent {\bf Future Work.} There are many avenues for future work. On the one hand, we would like to conduct systematic studies of other issues such as multiclass classification. We will also create a different series of artificial domains to study, more carefully, the impact of the data challenges considered here on Convolutional Networks. We also plan to extend our study to the domain characteristics responsible for adversarial examples, in a careful framework similar to the one used here. 
In a second avenue of research, we plan to study systematically the application of existing methods from both the classical machine learning community (under-sampling the majority class or over-sampling the minority class) and the Deep learning community (using cost-sensitive learning as well as regularisers) that address the data challenges discussed in this work.
Finally, the keen understanding of the characteristics that cause performance loss that we have gained here will be used to design tailored methods that can address these losses in the particular setting of deep learning.



\bibliographystyle{IEEEtran}
\bibliography{references}

\IEEEoverridecommandlockouts
\def\BibTeX{{\rm B\kern-.05em{\sc i\kern-.025em b}\kern-.08em
    T\kern-.1667em\lower.7ex\hbox{E}\kern-.125emX}}

\section{Supplemental Material}

\subsection{The equivalence of macro and weighted averages in the Balanced Setting experiments}
To formulate an explanation, considering $\tau^{+}$, $f^{+}$, $\tau^{-}$, and $f^{-}$ as true positive, false positive, true negative, and false negative we calculate the sensitivity, S, and specificity, Sp, for class 0 $\left(S^{0}, Sp^{0}\right)$ and class 1 $\left(S^{1}, Sp^{1}\right)$ as;
\begin{equation} \label{eq1}
    S^{0} = \frac{\tau^{-}}{\tau^{-}+f^{+}} S^{1} = \frac{\tau^{+}}{\tau^{+}+f^{-}}
\end{equation}

\begin{equation} \label{eq2}
    Sp^{0} = \frac{\tau^{+}}{\tau^{+}+f^{-}} Sp^{1} = \frac{\tau^{-}}{\tau^{-}+f^{+}}
\end{equation}

Based on the obtained expressions, the Geometric Mean for class 0 $\left(G^{0}\right)$ and class 1 $\left(G^{1}\right)$ can be mathematically defined as; 

\begin{equation} \label{eq3}
    G^{0}=\sqrt{\frac{\tau^{-}\cdot\tau^{+}}{\left(\tau^{-}+f^{+}\right)\cdot\left(\tau^{+}+f^{-}\right)}}
\end{equation}

\begin{equation} \label{eq4}
    G^{1}=\sqrt{\frac{\tau^{+}\cdot\tau^{-}}{\left(\tau^{+}+f^{-}\right)\cdot\left(\tau^{-}+f^{+}\right)}}
\end{equation}

Therefore, the macro average Geometric Mean can be written as;

\begin{equation} \label{eq5}
    G^{mac} = \frac{1}{2}\cdot G^{0}+\frac{1}{2}\cdot G^{1}
\end{equation}

The weighted average involves a calculation concentrating on the number of examples present in the training set for each class. If we consider the training set as X, then, $X_{0}$ represents the total number of instances that belong to class 0 and $X_{1}$ is that of class 1, such that, $|X|= X_{0}+ X_{1}$. The weighted average can then be denoted as;

\begin{equation} \label{eq6}
    G^{w} = \frac{X_{0}}{|X|}\cdot G^{0}+\frac{X_{1}}{|X|}\cdot G^{1}
\end{equation}

On careful analysis, we find that the expression $G^{0}$ and $G^{1}$ are similar. Therefore, $G^{mac}$ will be equal to $G^{0,1}$ such that $G^{0,1} = G^{0} = G^{1}$. Similarly, we can also conclude $G^{w}$ as $G^{0,1}$. Hence, we are getting $G^{mac} = G^{w}$.
\subsection{Stratified CV Experiments}

In this section, we present the performance of deep (with 5 hidden layers) and shallow (with 1 hidden layer) MLP models that we obtained after applying 10 Fold Stratified Cross Validation on the Backbone and Overlapped Domains. Figure \ref{fig:3} and Figure \ref{fig:4} illustrates the results obtained on the Backbone framework of different sizes whereas, Figure \ref{fig:17} shows the results obtained on the Overlapped Datasets.

\smallskip
\begin{figure}[t]
  \centering
  \begin{subfigure}{3cm}
    \centering\includegraphics[width=3cm]{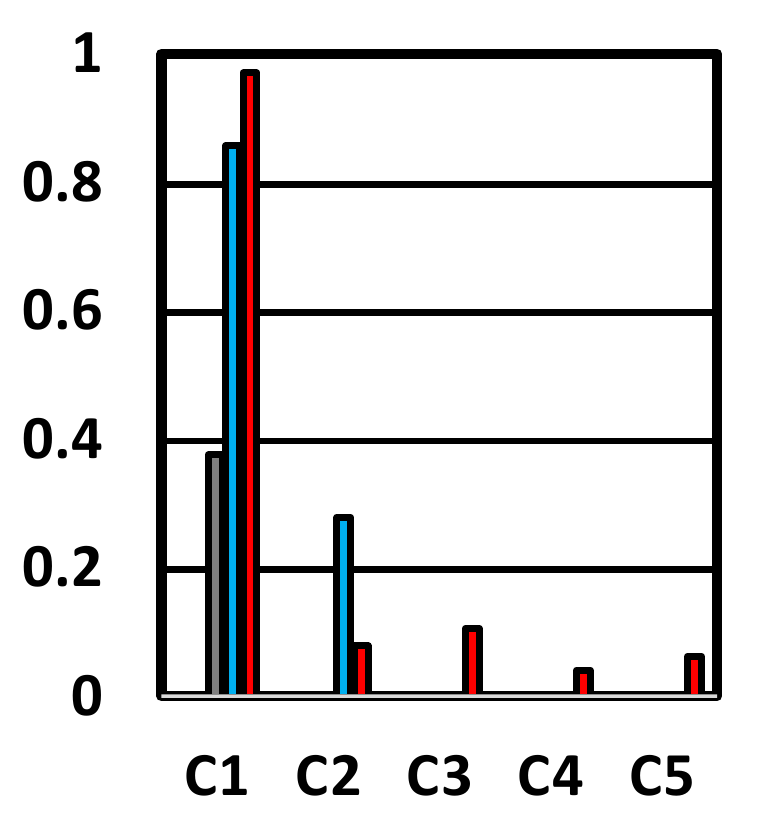}
    \caption{Model 1}
  \end{subfigure}
  \begin{subfigure}{3cm}
    \centering\includegraphics[width=3cm]{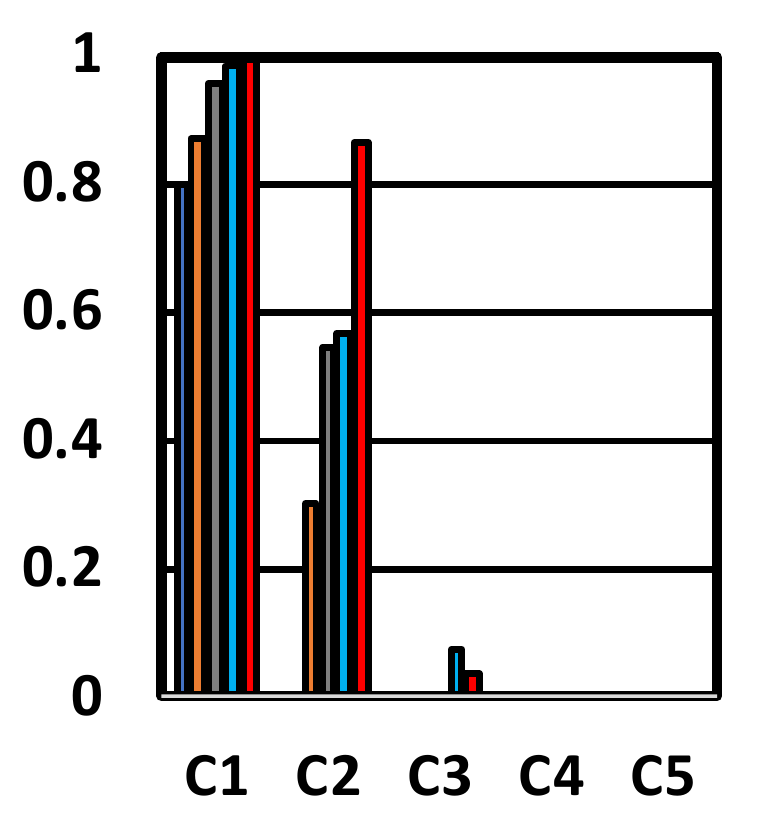}
    \caption{Model 5}
  \end{subfigure}
  \caption{MLP generated Macro G-Mean Scores by doing 10-fold Stratified Cross Validation for Size 1: (a) Model 1 and (b) Model 5.}
  \label{fig:3}
\end{figure}

\smallskip
\begin{figure}[t]
  \centering
  \begin{subfigure}{3cm}
    \centering\includegraphics[width=3cm]{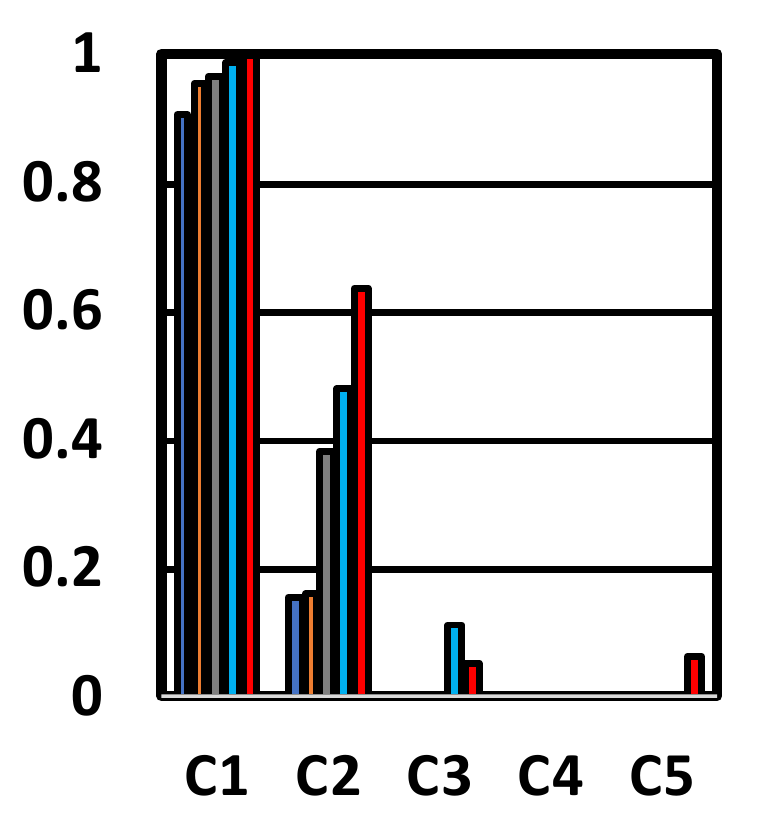}
    \caption{Model 1}
  \end{subfigure}
  \begin{subfigure}{3cm}
    \centering\includegraphics[width=3cm]{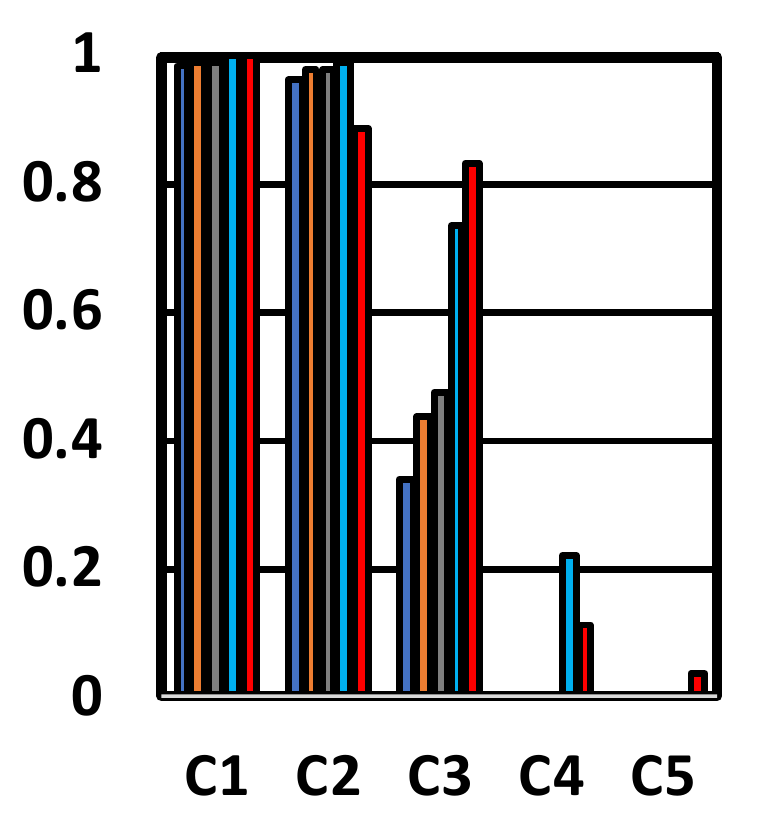}
    \caption{Model 5}
  \end{subfigure}
  \caption{MLP generated Macro G-Mean Scores by doing 10-fold Stratified Cross Validation for Size 5: (a) Model 1 and (b) Model 5.}
  \label{fig:4}
\end{figure}

\begin{figure*}[t]
  \centering
  \begin{subfigure}{8.5cm}
    \centering\includegraphics[height=3cm,width=8.5cm]{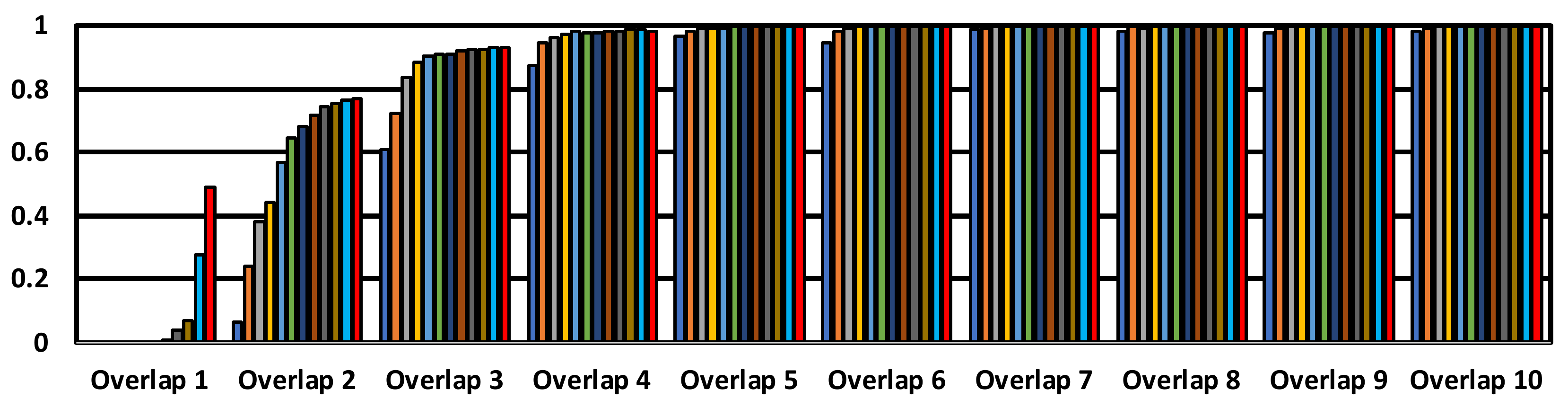}
    \caption{}
  \end{subfigure}
  \begin{subfigure}{8.5cm}
    \centering\includegraphics[height=3cm,width=8.5cm]{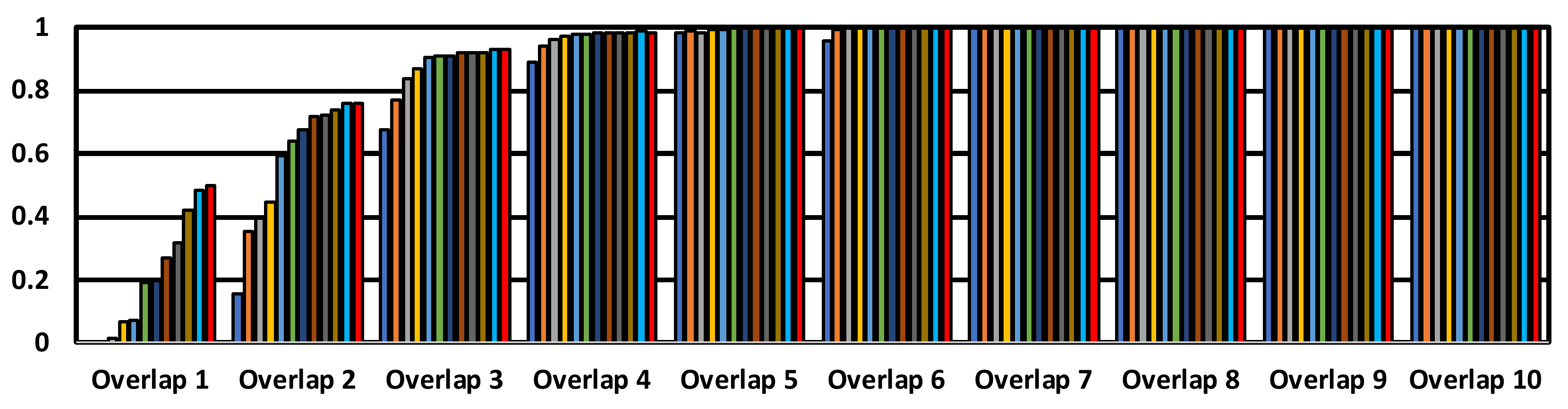}
    \caption{}
  \end{subfigure}
  \caption{MLP generated Macro G-Mean Scores by doing 10-Fold Stratified Cross Validation on the Overlapped Datasets: (a) Model 1 and (b) Model 5.}
  \label{fig:17}
\end{figure*}

\subsection{Results on Image Domains}

\begin{table}[htbp]
\caption{Distribution of rank of model performance on MNIST Fashion as a function of depth. These results indicated that on average there is a slight preference for shallower model on the highly imbalanced data.}
\begin{center}
\begin{tabular}{c|ccccc}
\hline
& \multicolumn{5}{c}{Sum of Rank}\\
Model Depth &	Balanced &	0.3 &	0.15 &	0.05 &	0.025 \\
\hline
\hline
1&12.0 &	15.0 &	13.0 &	21.0 &	13.0\\
2&9.0 &   10.0 &	9.0 &	17.0 &	10.0\\
3&7.0 &	6.0 &	14.0 &	9.0 &	12.0\\
4&18.0 &	16.0 &	12.0 &	13.0 &	17.0\\
5&22.0 &	18.0 &	20.0 &	9.0 &	22.0\\
\hline
\hline
\end{tabular}
\label{tab:ranksOfDepthOnMnistFashion}
\end{center}
\end{table}

\begin{figure*}[t]
 \centering
    \centering\includegraphics[width=4cm]{figures/mf3versus7.pdf}
    
    \centering\includegraphics[width=4cm]{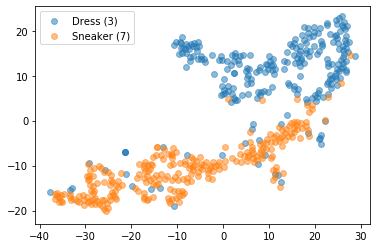}
    \centering\includegraphics[width=4cm]{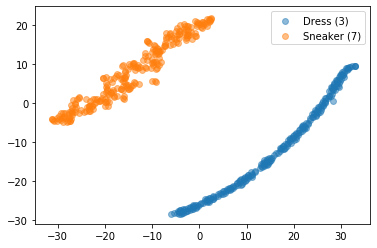}
    \centering\includegraphics[width=4cm]{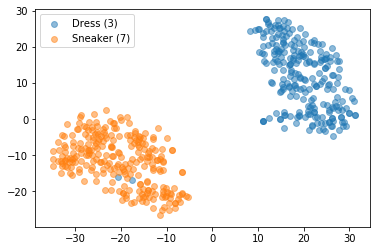}
    \centering\includegraphics[width=4cm]{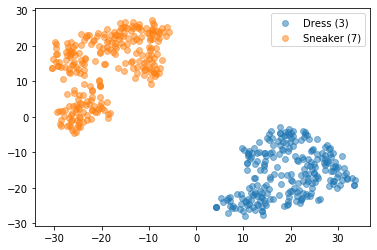}
    
    \centering\includegraphics[width=4cm]{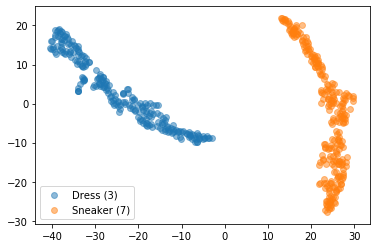}
    \centering\includegraphics[width=4cm]{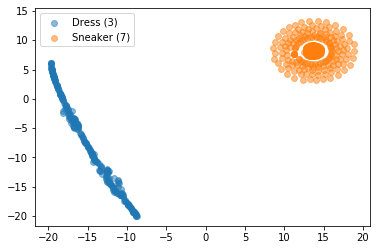}
    \centering\includegraphics[width=4cm]{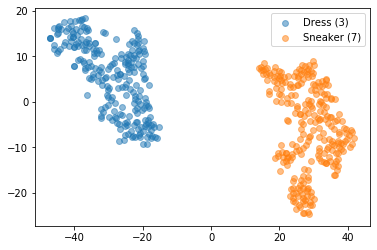}
    \centering\includegraphics[width=4cm]{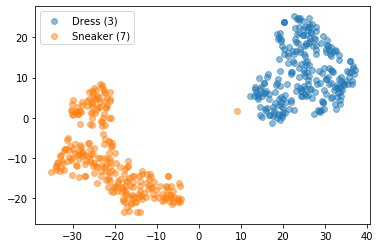}

    \centering\includegraphics[width=4cm]{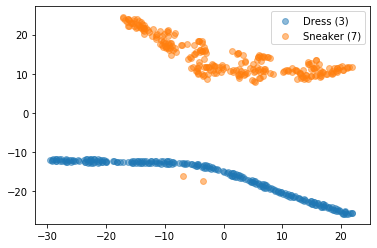}
    \centering\includegraphics[width=4cm]{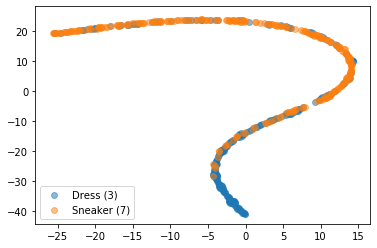}
    \centering\includegraphics[width=4cm]{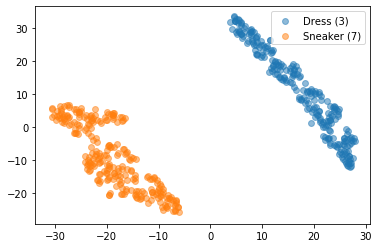}
                         \centering\includegraphics[width=4cm]{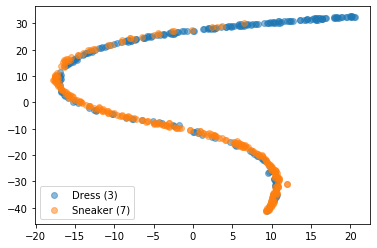}

 \caption{These plots illustrate the representation learned on the MNIST Fashion for classes 3 versus 7. The plot on the top row shows the T-SNE embedding of the original data. This reveals that this is a classification problem with low complexity due to the presence of a single linearly separable concepts. In the remaining rows, the first and second columns show the T-SNE plots for the representations learned after the last CNN block on the balanced (first column) and imbalanced (second column) data. The third and fourth columns show the T-SNE plots for the representations learned after the last fully connected layers for balanced (third column) and imbalanced (fourth column). Row-wise, the plot represent the results of CNNs with 1, 3, and 5 CNN blocks. These plots illustrate when data is imbalanced, increasing the number of CNN blocks to causes the classes to become overlapping in the learned representation. Thus, adding extra layers to an network training on imbalanced data can, in fact, make the learning problem more difficult.}
 \label{fig:cnnMnistFashion3_7Embeddings}
\end{figure*}

\end{document}